\documentclass{article}

\usepackage{microtype}
\usepackage{graphicx}
\usepackage{subcaption}
\usepackage{booktabs} % for professional tables

\usepackage{hyperref}

\usepackage{iclr2027_conference,times}
\usepackage{amsmath}
\usepackage{amssymb}
\usepackage{mathtools}
\usepackage{amsthm}
\usepackage{xspace}
\usepackage{wrapfig}
\usepackage{threeparttable}
\usepackage{multicol}
\usepackage{multirow}
\usepackage{subcaption}
\usepackage{algorithm}
\usepackage{algorithmic}
\usepackage[capitalize,noabbrev]{cleveref}

\theoremstyle{plain}
\newtheorem{theorem}{Theorem}[section]
\newtheorem{proposition}[theorem]{Proposition}

\theoremstyle{definition}
\newtheorem{definition}[theorem]{Definition}

\theoremstyle{remark}

\newcommand{\method}{Stylus\xspace}
\newcommand{\rev}[1]{{\color{black}{#1}}}
\usepackage{hyperref}
\usepackage[textsize=tiny]{todonotes}

\title{Stylization as Geodesic Navigation on Diffusion Manifolds}
\iclrfinalcopy
\small
\author{
Yan Li$^{\dagger\thanks{For any inquery, please reach out to:ylitz@connect.ust.hk}}$  \hspace{-7pt}
\And
Zehao Li$^{\ddagger}$ \hspace{-7pt}
\And
Xinlong Sun$^{\S}$  \hspace{-7pt}
\And
Fuchen Long$^{\S}$  \hspace{-7pt}
\And
Xi Chen$^{\S}$  \hspace{-7pt}
\And
Wenhan Luo$^{\dagger}$  \hspace{-7pt}
\And
Yike Guo$^{\dagger}$ 
\AND 
$^\dagger$~HKUST  \And $^{\ddagger}$ Peking University \And $^{\S}$ Tecent
}
\begin{document}




\maketitle


 % no special notice (required even if empty)

\vspace{-16pt}
\begin{abstract}
Facial stylization asks for a delicate transformation: the image should look unmistakably artistic, yet the person should remain unmistakably the same. Existing training-free and instruction-driven methods often fail to expose such control, either drifting away from identity or producing only weak stylization. We introduce \textbf{\method}, a training-free framework that treats facial style transfer as navigation on the curved latent manifold of a pretrained diffusion model. Rather than interpolating in flat Euclidean space, \method refines the interpolated content--style point by lowering a ratio-weighted discrete geodesic energy under the pullback metric, made tractable by a rank-one power-iteration approximation of the Jacobian. This principle drives two complementary controls: content--style latent interpolation for smooth semantic fusion, and self-attention query optimization for spatially adaptive style injection. The resulting dual-control mechanism provides continuous, fine-grained control over stylization strength while preserving identity-critical facial structure, all without additional training. Experiments show that \method achieves the strongest identity and content preservation at competitive stylization strength.

\end{abstract}

\section{Introduction}
\label{sec:intro}
Image stylization, which aims to transform images into artistic styles while preserving their original content characteristics~\cite{gatys2016image,kwon2022clipstyler,deng2022stytr2,zhang2023inversion,hong2023aespa,zhang2022domain,an2021artflow,yang2023zero}, has attracted significant research attention in recent years. While traditional style transfer methods have shown success with general images, facial image stylization presents unique challenges that require special consideration. While artistic style transfer is a well-established field, the specific challenge of stylizing facial images while maintaining identity has received surprisingly limited research attention in recent years~\cite{khowaja2024fistnet,yi2020unpaired,shi2019warpgan,chen2020animegan}. The key challenge of facial image stylization lies in striking a delicate balance - the stylization must maintain recognizable facial features that preserve the subject's identity while simultaneously achieving compelling artistic effects~\citep{cohen2025conditional}. This requires more precise control over the stylization process compared to general image stylization, making it a particularly demanding problem in the field.

Recent advances in diffusion models have demonstrated remarkable capabilities in various image generation and manipulation tasks~\cite{ho2020denoising,nichol2021improved,kim2022diffusionclip,wang2023not,kong2023leveraging}, with promising applications in style transfer~\cite{jeong2024training,wang2023stylediffusion,chung2024style,kwon2022diffusion}. More recently, large-scale DiT-based editing models~\cite{wu2025qwen,labs2025flux} have pushed the frontier of instruction-driven image manipulation and can be repurposed for stylization via natural-language prompts. Despite their impressive generative quality, several critical limitations persist. First, adapting conventional diffusion models to new artistic styles typically demands extensive computational resources and training time~\cite{zhang2023inversion}, as gradient estimation through large generative models remains costly despite recent advances~\citep{ren2026halforder,zou2026ovlr}. Second, existing diffusion-based stylization methods often lack precise control mechanisms over the stylization process, making it challenging to achieve desired artistic effects~\cite{chung2024style}. Third, they frequently struggle to maintain an optimal equilibrium between faithful style transfer and the preservation of essential identity features, which is particularly crucial for facial stylization. Finally, DiT-based editing models inherit a distinct set of weaknesses~\cite{wu2025qwen,labs2025flux}: they rely almost exclusively on textual prompts to specify the target appearance, which is inherently ambiguous for describing visual style; they provide no fine-grained knob to modulate stylization strength; and they tend to either over-edit, erasing the subject's identity, or under-edit, failing to faithfully transfer the reference style.

We trace these failures to a shared oversight: prior methods implicitly treat the diffusion latent space as \emph{flat Euclidean space}, where linear interpolation and uniform attention substitution are assumed to produce semantically valid results. In reality, diffusion models learn a \emph{curved manifold} on which valid images reside~\cite{kwon2022diffusion,park2023understanding}, and straight-line traversal of this manifold routinely crosses low-probability regions, yielding ghosting artifacts, identity drift, or weak stylization. Our key insight is to instead use the pullback geometry of this manifold to refine ratio-selected intermediate points toward locally low-energy geodesic points~\cite{shao2018riemannian}, exposing an interpolation parameter that continuously modulates stylization strength.
%Guided by this principle, we propose \textbf{\method}, a training-free framework that realizes geodesic navigation at two complementary stages of the diffusion process. First, a \emph{linear style interpolation} mechanism computes geodesic paths between content and style latents, and samples along this mathematically optimal trajectory to achieve smooth, semantically consistent content–style fusion. Second, a \emph{dynamic style injection control} mechanism extends attention-based stylization~\cite{chung2024style} by computing geodesic paths between content and style queries in the attention space; sampling along these paths across diffusion steps enables spatially adaptive injection: stronger in regions that benefit from artistic expression and gentler in identity-critical facial features while recovering fine-grained details that latent-space interpolation alone cannot capture. Together, these two mechanisms resolve the flexibility-quality trade-off without any model training.

Guided by this principle, we propose \textbf{\method}, a training-free framework that performs geodesic-guided point refinement at two complementary stages. First, \emph{linear style interpolation} refines a ratio-selected content--style latent by minimizing a ratio-weighted discrete energy. Second, \emph{dynamic style injection control} applies the same refinement to a ratio-selected point between content and style queries in the attention space, yielding spatially adaptive injection while recovering fine-grained details that latent-space interpolation alone cannot capture.
%To address these challenges, we propose a novel training-free approach \textbf{\method} that offers precise control over the face stylization process. First, we develop a linear style interpolation mechanism that leverages geodesic paths~\cite{shao2018riemannian}, the shortest paths between two points on a curved surface, to enable careful control over the initial content-style fusion. By computing the geodesic path between the content and style images in the latent space, we can sample different points along this mathematically optimal trajectory, providing fine-grained control over style integration while ensuring smooth and natural transitions. Second, we implement a dynamic style injection control mechanism that adaptively modulates the balance between content and style features during the diffusion process. Building upon attention-based style injection~\cite{chung2024style}, we compute a geodesic path between the content query and style query in the attention space. By sampling points along this path at different stages of diffusion, we can precisely control how much style information is injected into the content features. This dynamic control allows for stronger style injection in areas that benefit from artistic expression while maintaining more conservative injection rates in regions crucial for identity preservation, such as facial features.

The main contributions of our work can be summarized as follows: 
\vspace{-8pt}
\begin{itemize}\setlength{\itemsep}{-1pt}
    \item A novel training-free method for facial stylization that can be generalized across various stable diffusion architectures, achieving state-of-the-art performance while providing fine-grained control over the stylization process.
    \item A dual-control framework combining geodesic-based style interpolation and adaptive style injection, enabling precise content-style fusion through optimal trajectories in latent space while dynamically optimizing the balance between stylization and identity preservation.
    %via attention-based geodesic sampling.
   % \item A geodesic-based style interpolation mechanism that enables precise control over content-style fusion in the latent space through mathematically optimal trajectories
   % \item An adaptive style injection control system that dynamically optimizes the balance between artistic stylization and identity preservation through attention-based geodesic sampling in the diffusion process
    \item Extensive empirical evaluation and rigorous experimental analysis demonstrating significant performance improvements across a comprehensive suite of quantitative metrics and systematic qualitative assessments.
\end{itemize}

\section{Related Work} 

\subsection{Image Style Transfer}
Image style transfer has been an active research area in computer vision. Early approaches like Gatys et al.~\cite{gatys2016image} pioneered neural style transfer by using CNN features to separate and recombine content and style. Since then, numerous methods have been proposed to improve style transfer quality and efficiency~\cite{johnson2016perceptual, huang2017arbitrary}. Recent transformer-based methods like StyTr2~\cite{deng2022stytr2} leverage self-attention mechanisms for better feature transformation. CLIPStyler~\cite{kwon2022clipstyler} enables text-guided style transfer by utilizing CLIP's multimodal representations. With the advent of diffusion models~\cite{nichol2021improved}, several works have explored their potential for style transfer. Wang et al.~\cite{wang2023stylediffusion} proposed StyleDiffusion for controllable disentangled style transfer. Zhang et al.~\cite{zhang2023inversion} developed an inversion-based approach for more precise style control. Domain-aware methods like~\cite{zhang2022domain} incorporate contrastive learning to better capture domain-specific style features.
Recent works have also focused on improving artistic quality while maintaining content fidelity~\cite{liu2025samam,zhang2025hsi,wang2025omnistyle,chittersu2025insert}. AesPA-Net~\cite{hong2023aespa} introduces aesthetic pattern awareness for more visually pleasing results. ArtFlow~\cite{an2021artflow} employs reversible neural flows to achieve unbiased style transfer. Yang et al.~\cite{yang2023zero} proposed a zero-shot contrastive loss for text-guided diffusion style transfer, demonstrating the potential of diffusion models in this domain. More recently, DiT-based image editing models, such as Qwen-Image-Edit~\cite{wu2025qwen} and FLUX.2-dev~\cite{labs2025flux}, have shown strong editing capabilities, and can be adapted to stylization through textual prompts. However, their reliance on language prompts makes visual style specification ambiguous, and they provide limited fine-grained control over stylization strength and identity preservation.

\subsection{Facial Image Stylization}

Recognizing the unique challenges of facial stylization, researchers have developed specialized approaches for face images~\cite{khowaja2024fistnet}. Early attempts adapted general style transfer methods with face-specific constraints~\cite{selim2016painting}. More recent works have incorporated facial landmark detection~\cite{zhang2017style} and identity preservation mechanisms~\cite{chung2024style}.
StyleGAN-based approaches~\cite{karras2019style,karras2020analyzing} and unsupervised image-to-image translation methods have shown promise in facial stylization tasks like selfie-to-anime conversion. However, these methods typically require task-specific training to achieve satisfactory results. While many existing approaches excel at single-style transfer, they often struggle with multi-style fusion and customized style combinations. Methods like DreamBooth~\cite{ruiz2022dreambooth} and Textual Inversion~\cite{gal2022image} face similar challenges when attempting to combine multiple artistic styles like cartoon, anime, and arcane aesthetics. A key limitation across these approaches is their difficulty in preserving facial identity features during style transfer. The misalignment between source and target domain facial features frequently results in artifacts that compromise the quality of the stylized output.
%Notable approaches include StyleGAN-based methods~\cite{richardson2021encoding,yang2022pastiche,yang2022vtoonify} that leverage the style space of pre-trained GANs, and attention-based frameworks like StyTR$^2$~\cite{deng2022stytr2} that better preserve facial structures. While these methods show improved stylization performance, they often require extensive training or lack fine-grained control over the stylization process.

Our work advances the field with a training-free dual-control framework that enables fine-grained control over both initial style fusion and ongoing style injection in diffusion models, achieving precise stylization while preserving facial identity.

\section{Methodology}

\subsection{Theoretical Foundation}\label{sec4.1}
%Basically, we want to transfer the problem in coordinate space $\mathcal{X}$ to a Riemannian manifold $\mathcal{M}$ through the mapping $g$. For a general background on geodesic path computation, see Appendix~\ref{sec:geodesic}.  Before introducing the geodesic-guided style interpolation in Section \ref{sec:style_interp}, we state two enabling results that follow from the pullback construction and certify that the surrogate energy computed from $g(\cdot)$ locally matches the energy on $\mathcal{M}$. First, we define the corresponding Riemannian structures and geodesic distances in $\mathcal{X}$ and $\mathcal{M}$.
Let the frozen U-Net be decomposed as
\(\epsilon_\theta(x,t,c)=g_u(g_d(x,t,c),t,c)\). Throughout this section,
\(g(x):=g_d(x,t,c)\in\mathbb R^p\) denotes the vectorized down-path/bottleneck
hidden state at fixed timestep \(t\) and conditioning \(c\), while
\(x\in\mathcal X=\mathbb R^n\) is the vectorized diffusion latent. In our
setting \(p>n\); \(g\) is not the final noise prediction. Assuming that
\(J_g\) has full column rank in the local region visited by refinement,
\(g\) is an immersion and \(G=J_g^\top J_g\) defines a pullback Riemannian
metric on that region. This gives a local interpretation of distances after
the hidden-state map; it is not a global claim about the complete diffusion
trajectory.
%For a general background on geodesic path computation, see Appendix~\ref{sec:geodesic}. Before introducing the geodesic-guided style interpolation in Section \ref{sec:style_interp}, we have the theoretical results that guarantee the feasibility of our method. First, we will define and analyze the corresponding Riemannian structures and geodesic distances in $\mathcal{X}$ and $\mathcal{M}$.
\begin{definition}
    Let $\gamma:[0,1]\to\mathcal X$ be an absolutely continuous (AC) curve. 
The arc length of $\gamma$ with respect to the pullback Riemannian metric $\mathsf g$ is
\begin{equation*}
\begin{aligned}
L_{(\mathcal X,\mathsf g)}(\gamma)
\;=\;
\int_0^1 \big\|\dot\gamma(t)\big\|_{ g,\gamma(t)}\,dt
\;=\; 
\int_0^1 \sqrt{\,\dot\gamma(t)^{\top} G\!\big(\gamma(t)\big)\,\dot\gamma(t)\,}\,dt.
\end{aligned}    
\end{equation*}

Let $\eta(t)=g(\gamma(t))\in \mathcal M$. 
The arc length of $\eta$ with respect to the metric induced by the ambient Euclidean inner product is
\[
\small
\begin{aligned}
L_{(\mathcal M,\|\cdot\|_2)}(\eta)
\;=\;
\int_0^1 \big\|\dot\eta(t)\big\|_2\,dt
\;=\;
\int_0^1 \Big\|\,\tfrac{d}{dt}\big(g\circ\gamma\big)(t)\,\Big\|_2\,dt.
\end{aligned}
\]
\noindent\textbf{Geodesic distances.}
For $x_1,x_2\in\mathcal X$, the geodesic distance induced by $\mathsf g$ is
\[
\begin{aligned}
d_{\mathsf g}(x_1,x_2)
\;:=\;
\inf\Big\{\,L_{(\mathcal X,\mathsf g)}(\gamma)\;:\;
\gamma\in AC([0,1],\mathcal X),  \gamma(0)=x_1,\ \gamma(1)=x_2\Big\}.
\end{aligned}
\]
 For $y_1,y_2\in\mathcal M$, the geodesic distance induced by the Euclidean metric restricted to $\mathcal M$ is
\[
\begin{aligned}
d_{\mathcal M}(y_1,y_2)
\;:=\;
\inf\Big\{\,L_{(\mathcal M,\|\cdot\|_2)}(\eta)\;:\;
\eta\in AC([0,1],\mathcal M), \eta(0)=y_1,\ \eta(1)=y_2\Big\}.
\end{aligned}
\]
\end{definition}
The following proposition exhibits the local isometry between $(\mathcal X,\mathsf g)$ and $(\mathcal M,\|\cdot\|_2)$. The proof can be found in Appendix~\ref{sec:geodesic}. 
\begin{proposition}[Local isometry of $(\mathcal X,\mathsf g)$ into $(\mathcal M,\|\cdot\|_2)$]
Let $g:\mathcal X\to\mathbb R^p$ be continuously differentiable and assume $J_g(x)$ has full column rank at $x\in\mathcal X$.
Consequently, there exists a neighborhood $U\ni x$ such that $g|_U:U\to \mathcal M_U$ is a Riemannian isometry onto its image (lengths of curves are preserved), i.e.,
\[
\begin{aligned}
 L_{(\mathcal X,\mathsf g)}(\gamma)=  L_{(\mathcal M_U,\|\cdot\|_2)}(g\circ\gamma)
\quad    \text{for all absolutely continuous }
 \gamma:[0,1]\to U.
\end{aligned}
\]
\end{proposition}
\begin{proof}
By definition of the pullback metric,
$\mathsf g_x(u,v)=\langle J_g(x)u,\,J_g(x)v\rangle_2$,
so $J_g(x)$ preserves inner products on tangent spaces.
%Full rank implies $J_g(x)$ is injective; by the inverse function theorem there is a neighborhood $U$ of $x$ on which $g$ is a $C^1$ diffeomorphism onto its image $\mathcal M_U$.
Since full column rank is an open condition, $J_g$ has constant rank $n$ on a sufficiently small neighborhood $U$ of $x$. By the constant rank theorem, after possibly shrinking $U$, the restriction $g|_U$ is a $C^1$ embedding and hence a diffeomorphism onto its image $\mathcal M_U$.
For any absolutely continuous $\gamma:[0,1]\to U$,
\[
\big\|\dot\gamma(t)\big\|_{g,\gamma(t)}
=\big\|J_{g}(\gamma(t))\dot\gamma(t)\|_2
=\big\|\tfrac{d}{dt}\big(g\circ\gamma\big)(t)\big\|_2 ,
\]
hence lengths are preserved after integration. This is precisely the definition of a local Riemannian isometry onto $\mathcal M_U$ endowed with the Euclidean induced metric.
\end{proof}

Within $U$, optimizing arc length or taking Riemannian gradients in $\mathcal X$ under the pullback metric is thus identical to doing so on $\mathcal M$. Practically, the chord length $\|g(x_{k+1})-g(x_k)\|_2$ serves as a surrogate for the short geodesic arc length between nearby points; this is a local, per-step approximation, valid for small steps and moderate curvature.
%The local isometry proposition ensures \emph{geometric fidelity}: optimizing arc length, tracing geodesics, or taking Riemannian gradients in latent space $\mathcal X$ with the pullback metric is  identical to doing so on the generated manifold $\mathcal M$. Practically, it justifies  the straight-line length $\|g(x_{k+1})-g(x_k)\|_2$ between two points in the embedding space $\mathcal{M}$ can serve as a surrogate for the short geodesic arc length between them. As long as the step size is small and the curvature is not too large, this chord length is an accurate local approximation to the geodesic length. 
%The following proposition upgrades the local statement to an end-to-end guarantee: on any open set $U$ where 
%$g|_{U}$ is a diffeomorphism onto its image, the geodesic distance between endpoints computed in latent space $(\mathcal{X},\mathsf g)$ exactly matches the geodesic distance between their images on the generated manifold $\mathcal{M}$. Consequently, shortest paths, interpolation costs, and energy minima are identical whether you pose the problem in $\mathcal{X}$ with the pullback metric or directly on $\mathcal{M}$ with the induced metric. 
The following proposition gives the corresponding distance equality on $U$. Since it requires both points to lie in one such neighborhood, we use it only locally, around the current iterate $x_m$, not as a global content-to-style guarantee.
\begin{proposition}[Local distance equality between $(U,\mathsf g)$ and $\mathcal M_U$] Endow $U$ with the pullback Riemannian metric $\mathsf{g}_x(u,v):=\langle J_g(x)u,J_g(x)v\rangle$ and $\mathcal M_U$ with the Euclidean induced metric.
Then for all $x_1,x_2\in U$,
\[
d_{\mathsf g,U}(x_1,x_2)\;=\;d_{\mathcal M_U}\big(g(x_1),g(x_2)\big),
\]
where both distances are defined using curves restricted to $U$ and $\mathcal M_U$, respectively.
\end{proposition}
\iffalse
\begin{proposition}[Distance equality between $(\mathcal X,\mathsf g)$ and $(\mathcal M,d_{\mathcal M})$] Endow $\mathcal X$ with the pullback Riemannian metric $\mathsf{g}_x(u,v):=\langle J_g(x)u,J_g(x)v\rangle$ and $\mathcal M_U$ with the Euclidean induced metric (hence geodesic distance $d_{\mathcal M}$).
Then for all $x_1,x_2\in U$,
\[
d_\mathsf g(x_1,x_2)\;=\;d_{\mathcal M}\big(g(x_1),g(x_2)\big),
\]
where $d_\mathsf g$ is the geodesic distance of $(\mathcal X,\mathsf g)$ and $d_{\mathcal M}$ is the geodesic distance on $(\mathcal M_U,\langle\cdot,\cdot\rangle_2)$.
\end{proposition}
\fi 
\begin{proof}
For any absolutely continuous curve $\gamma:[0,1]\to U$ we have, by definition of the pullback metric,
\[
\small
\begin{aligned}
\big\|\dot\gamma(t)\big\|_{g,\gamma(t)} 
=\sqrt{\dot\gamma(t)^\top J_g(\gamma(t))^\top J_g(\gamma(t))\,\dot\gamma(t)}  
=\big\|\tfrac{d}{dt}\big(g\circ\gamma\big)(t)\big\|_2 .
\end{aligned}
\]
Since length is preserved pointwise and hence after integration, we have $L_{(U,\mathsf g)}(\gamma)=L_{(\mathcal M_U,\|\cdot\|_2)}(g\circ\gamma)$. Taking the infimum over all such $\gamma$ in $U$ with $\gamma(0)=x_1,\gamma(1)=x_2$ yields
\[
d_{\mathsf g,U}(x_1,x_2)\;\ge\; d_{\mathcal M_U}(g(x_1),g(x_2)),
\]
because any competitor in $U$ maps to a competitor in $\mathcal M_U$ with the same length.
%Since length is preserved pointwise and hence after integration, we have $L_{(\mathcal X,\mathsf g)}(\gamma)=L_{(\mathcal M_U,\|\cdot\|_2)}(g\circ\gamma)$. Taking infimum over all $\gamma$ with $\gamma(0)=x_1,\gamma(1)=x_2$ yields
%\[
%d_\mathsf g(x_1,x_2)\;\ge\; d_{\mathcal M}(g(x_1),g(x_2)),
%\]
%because any competitor in $\mathcal X$ maps to a competitor in $\mathcal M_U$ with the same length.

Conversely, since $g|_U$ is a diffeomorphism, every absolutely continuous $\eta:[0,1]\to\mathcal M_U$ with $\eta(0)=g(x_1)$, $\eta(1)=g(x_2)$ lifts uniquely to $\gamma=(g|_U)^{-1}\circ\eta$ in $U$, and the same length identity shows $L_{(\mathcal X,\mathsf g)}(\gamma)=L_{(\mathcal M_U,\|\cdot\|_2)}(\eta)$. Taking infimum over such $\eta$ gives the reverse inequality and hence equality of distances.
\end{proof}
The above results explain why our interpolation objective is defined after the
latent points are passed through $g$. A small displacement in the raw latent
space may have very different effects after the generator mapping, and the
pullback metric measures such effects by the change in $g(x)$. %Therefore, in the next section, we refine the intermediate point $x_m$ by minimizing a discrete energy built from the mapped representations $g(x_c)$, $g(x_m)$, and $g(x_s)$.}

\subsection{Geodesic-Guided Style Interpolation} \label{sec:style_interp}
In the context of style transfer, we aim to find an optimal path between content and style representations. Let us consider two points $x_c, x_s \in \mathcal{X}$ representing the content and style images in the coordinate space $\mathcal{X}$. These points can be connected by a continuous curve $\gamma: [0,1] \rightarrow \mathcal{X}$ with $\gamma(0) = x_c$ and $\gamma(1) = x_s$. Following Section~\ref{sec:geodesic}, when this curve is mapped through the hidden-state map $g$, it creates a trajectory $\gamma_g: [0,1] \rightarrow \mathcal{M}$ on the Riemannian manifold $\mathcal{M}$, where $\gamma_g = g \circ \gamma$. Each point along this manifold curve represents a unique blend of content and style characteristics, enabling controlled style transfer.

%To begin finding this optimal path, we start with a simple linear interpolation $x_m = \delta x_s + (1-\delta) x_c$, where $\delta \in [0,1]$ controls the mixing ratio between content and style. While this linear interpolation provides an initial point in the coordinate space $\mathcal{X}$, our goal is to refine it into a more geometrically meaningful path that better preserves both content and style characteristics when mapped onto the manifold $\mathcal{M}$ through the generative model $g$.
We start with a simple linear interpolation $x_m = \delta x_c + (1-\delta) x_s$, where $\delta \in [0,1]$ controls the mixing ratio between content and style (larger $\delta$ keeps the initial point closer to the content). While this linear interpolation provides an initial point in the coordinate space $\mathcal{X}$, our goal is to refine it into a more geometrically meaningful point that better preserves both content and style characteristics when mapped onto the manifold $\mathcal{M}$ through the hidden-state map $g$.

The curve $\gamma$ follows a direction from the content point $x_c$ to the style point $x_s$, and this directional path is preserved when mapped to the corresponding curve on the Riemannian manifold $\mathcal{M}$ through the hidden-state map $g$.
By performing forward finite difference, we can obtain the approximate velocity of the curve $\gamma$ at point $x_m$, which under our initialization lies at parameter $1-\delta$ along the path:
\begin{equation}
    v_m = (g(x_s) - g(x_m))/\delta.
\end{equation}
%By performing forward finite difference, we can obtain the approximate velocity of the curve $\gamma$ at point $x_m$:
%\begin{equation}
%    v_m = (g(x_s) - g(x_m))/(1-\delta).
%\end{equation}
%To quantify the optimality of this path, we derive a discrete version of the energy function from Equation~\ref{eq:energy} (refer to Appendix~\ref{sec:geodesic}). 
To quantify the optimality of this point, we use the discrete two-chord version of the energy function in Equation~\ref{eq:energy} (the derivation is given in Appendix~\ref{sec:geodesic}):
\begin{equation}\label{eq5}
\small
    E \approx \alpha\left(\delta||g(x_m)-g(x_c)||^2
    +(1-\delta)||g(x_s)-g(x_m)||^2\right),
\end{equation}
%where $\alpha $ is a normalization factor that accounts for the relative distances between points. By our theoretical results in Section \ref{sec4.1}, the geodesic distance in space $\mathcal{X}$ is equivalent to the $\Vert\cdot\Vert_2$ norm in space $\mathcal{M}$ by the mapping $g$, which aligns with the insight in Equation \ref{eq5}. 
where $\alpha$ is a constant scale factor. By Section \ref{sec4.1}, the chord lengths in $\mathcal{M}$ locally match geodesic lengths in $\mathcal{X}$, which motivates Equation \ref{eq5}. The ratio-dependent weights make the unconstrained mapped-space target $\delta g(x_c)+(1-\delta)g(x_s)$, consistent with the initialization; if this target lies outside $g(\mathcal{X})$, the full objective instead seeks a local projection onto $g(\mathcal{X})$.

\begin{wrapfigure}{r}{0.5\textwidth}
    \centering
    \vspace{-12pt}
    \includegraphics[width=\linewidth]{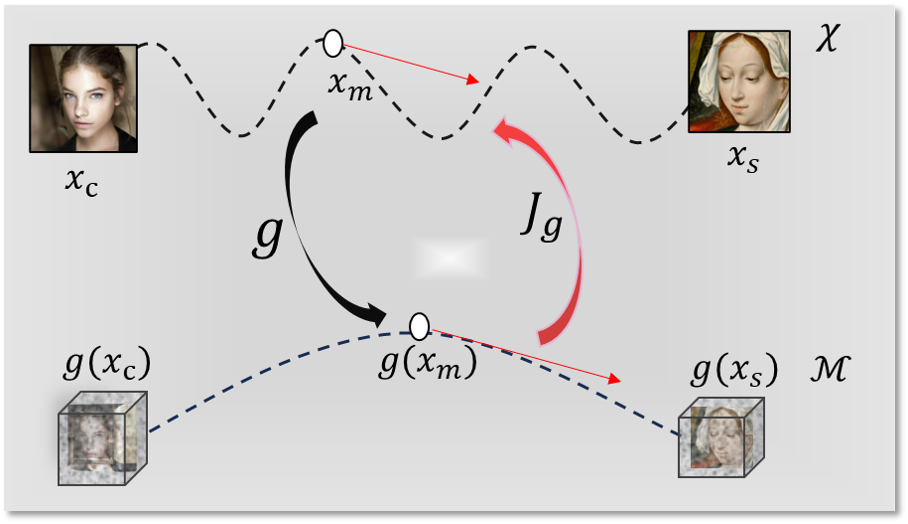}
    \caption{Geometric interpretation of the gradient computation: the hidden-state map $g$ maps points from coordinate space $\mathcal{X}$ to the manifold $\mathcal{M}$, while the Jacobian $J_g$ projects the manifold gradient onto the tangent space of $\mathcal{X}$.}
    \label{fig:jocabian_projection}
    \vspace{-12pt}
\end{wrapfigure}

%To get the geodesic path with fixed start point and end point as $x_c$ and $x_s$, we need to minimize the discrete energy function. One way is to perform gradient descent on $x_m$, the gradient on the point $x_m$ is:
To refine $x_m$ with fixed endpoints $x_c$ and $x_s$, we minimize the discrete energy function by gradient descent on $x_m$. The gradient on the point $x_m$ is:
\begin{equation}
     \nabla_{x_m}E = -2\alpha J_g^{\top}(x_m)
    \left(\delta g(x_c)+(1-\delta)g(x_s)-g(x_m)\right).
    \label{eq:grad}
\end{equation}
The gradient in Equation~\ref{eq:grad} forms a finite-difference residual in $\mathcal{M}$ and pulls it back to $\mathcal{X}$ through $J_g^{\top}$, which removes its component normal to the manifold (Figure~\ref{fig:jocabian_projection}). 
%The gradient in Equation~\ref{eq:grad} consists of two key components: a finite-difference second derivative on the manifold $\mathcal{M}$ and the Jacobian of $g$ derived through the chain rule. The finite-difference term in $\mathcal{M}$ space contains both tangential and normal components relative to the manifold's tangent space. The Jacobian $J_g$ serves to project out the normal component, effectively mapping the gradient from the manifold $\mathcal{M}$ back to the tangent space of $\mathcal{X}$, as illustrated in Figure~\ref{fig:jocabian_projection}. 
\textcolor{black}{This update is another direct consequence of the pullback formulation in
Section~\ref{sec4.1}: the finite-difference residual is first formed in the
mapped space $\mathcal M$, and the transpose Jacobian $J_g^\top$ pulls this
descent signal back to the coordinate space $\mathcal X$.} Through this projection mechanism, we can perform gradient descent directly on $x_m$ in the coordinate space $\mathcal{X}$ to find the geodesic path while respecting the manifold's geometric structure.

\subsection{Efficient Jacobian Approximation}
\textcolor{black}{The gradient in Equation~\eqref{eq:grad} shows how the pullback geometry enters
the actual update: the descent signal is computed in the mapped space and then
pulled back through $J_g^\top$. Thus, the practical computation only requires an
efficient approximation of this Jacobian-induced direction.}
However, computing the Jacobian of a generative model is computationally expensive, particularly for diffusion-based models. To address this challenge, we approximate the singular value decomposition (SVD) of the Jacobian using power iteration~\cite{haas2024discovering} over the matrix $J_g^\top J_g$, whose eigenvectors correspond to the right singular vectors of $J_g$. 

In our implementation, $g(x_m)=g_d(x_m,t,c)$ is the fixed-timestep down-path/bottleneck hidden-state map defined in Section~\ref{sec4.1}. Its Jacobian and SVD are:
\begin{equation}
    J_g(x_m) = \frac{\partial g_d(x_m,t,c)}{\partial x_m}
    = U\Sigma V^{\top} = \sum_i \sigma_i u_i v_i^{\top},
\end{equation}
where $v_1$ is the leading right singular vector. We begin with a random vector $v$; the product $J_g^{\top} J_g v$ can be computed as:
\begin{equation}
    J_g^{\top} J_g v = \frac{\partial}{\partial x_m}
    \big\langle g_d(x_m,t,c),\, \mathrm{sg}(J_g v)\big\rangle
\label{eq:singular_vec}
\end{equation}
and
\begin{equation}
    J_g v = \frac{\partial}{\partial a}
    g_d(x_m+a v,t,c)\Big|_{a=0},
\end{equation}
where $\mathrm{sg}(\cdot)$ denotes stop-gradient.
Iterating Equation~\ref{eq:singular_vec} with normalization converges to $v_1$ at rate $O\big((\sigma_2^2/\sigma_1^2)^K\big)$ after $K$ iterations.

We then replace the exact pullback direction $J_g^{\top} r_m=\sum_i \sigma_i (u_i^{\top} r_m)\,v_i$, with $r_m := \delta g(x_c)+(1-\delta)g(x_s)-g(x_m)$, by its leading term:
\begin{equation}
    \eta_m = -2\alpha\, v\, (J_g v)^{\top}
    \big(\delta g(x_c)+(1-\delta)g(x_s)-g(x_m)\big),
    \label{eq:rank1_update}
\end{equation}
where $(J_g v)^{\top} r_m$ is a scalar. This rank-one surrogate drops the subdominant directions $v_{i>1}$, in the spirit of efficient gradient estimation~\citep{li2024beyond,li2026optimal}; a top-$k$ comparison (Appendix~\ref{app:topk}) shows that adding them changes the results only marginally at a substantially higher cost. We update $x_m \leftarrow x_m - \lambda \eta_m$ with step size $\lambda$ and stop once the energy change drops below $10^{-4}$ (within 5--10 iterations in practice).
\subsection{Style Injection Control} \label{sec:style_injection_control}
Interpolating latent codes as in Section~\ref{sec:style_interp} can degrade content details. To better preserve content, prior work~\cite{chung2024style} injects style via self-attention (SA) layers, using query features from the content image and key/value features from the style image. However, since query features largely determine spatial structure, using only the content query weakens the applied style.
We propose to find an optimal interpolation between the content query ($Q_c$) and style query ($Q_s$). A naive linear interpolation, $Q_m = (1-\beta) Q_c + \beta Q_s$, often produces ghosting artifacts. Instead, we find a geodesic path in the query feature space by adapting the energy minimization from Section~\ref{sec:style_interp}. We initialize $Q_m$ with linear interpolation and then minimize the ratio-weighted energy
\[
E_\beta=(1-\beta)\|f(Q_m)-f(Q_c)\|^2
        +\beta\|f(Q_s)-f(Q_m)\|^2.
\]
Its gradient is:
\begin{equation}
    \nabla_{Q_m} E_\beta = -2J_{f}^{\top}
    \big((1-\beta)f(Q_c)+\beta f(Q_s)-f(Q_m)\big),
\end{equation}
where $f$ is the attention module. The key and value features are held fixed. As in the latent stage, the unconstrained mapped-space target is $(1-\beta)f(Q_c)+\beta f(Q_s)$, and the truncated refinement is a small correction of the linear query interpolant. This query-space refinement follows the same pullback-energy principle as
Section~\ref{sec4.1}, with the hidden-state map $g$ replaced by the attention
module $f$. In other words, the interpolation between $Q_c$ and $Q_s$ is
measured after the queries are transformed by $f$, rather than by their raw
Euclidean distance in query space.

We approximate the Jacobian $J_f$ using power iteration as before. The vector-Jacobian products are computed via:
\begin{equation}
    J_f^{\top}J_f v = \frac{\partial}{\partial Q_m}(f(Q_m), J_f v)
\end{equation}
and 
\begin{equation}
    J_f v = \frac{\partial}{\partial a}f(Q_m + av)|_{a=0},
\end{equation}
The update uses the same rank-one surrogate as Equation~\ref{eq:rank1_update}. By performing gradient descent on $Q_m$, we obtain modified query features that preserve content structure while faithfully representing the desired style.

\section{Experiment}
\subsection{Experiment Setting}
\textbf{Implementation:}  We use the official pretrained Stable Diffusion models (SD1.4, SD1.5, SD2.0, SD2.1), and the sample step is set to 50, the guidance scale is set to 3.0. The parameter $\delta$ is set to 0.9, $\alpha$ is set to 5, $\beta$ is set to 0.25 in our method.  All experiments are conducted on a single NVIDIA H800 GPU. The max iteration for geodesic optimization is set to 10, with early stopping at an energy change below $10^{-4}$. The implementation details can refer to the Appendix~\ref{app:implement} and Appendix~\ref{app:compute_cost}. 
%We use the official pretrained Stable Diffusion models (SD1.4, SD1.5, SD2.0, SD2.1), and the sample step is set to 50, the guidance scale is set to 3.0. The parameter $\delta$ is set to 0.9, $\alpha$ is set to 5, $\beta$ is set to 0.25 in our method.  The experiment is conducted on a single NVIDIA H800 GPU. The max iteration for geodesic optimization is set to 10. For more details, please refer to the Appendix~\ref{app:implement} and Appendix~\ref{app:compute_cost}. 

%For more details, please refer to the Appendix~\ref{app:implement}.
%For the detailed experiment settings and implementations, please refer to Appendix~\ref{app:implement}.
\begin{figure*}[!h]
    \centering
    \vspace{-8pt}
    \includegraphics[width=\linewidth]{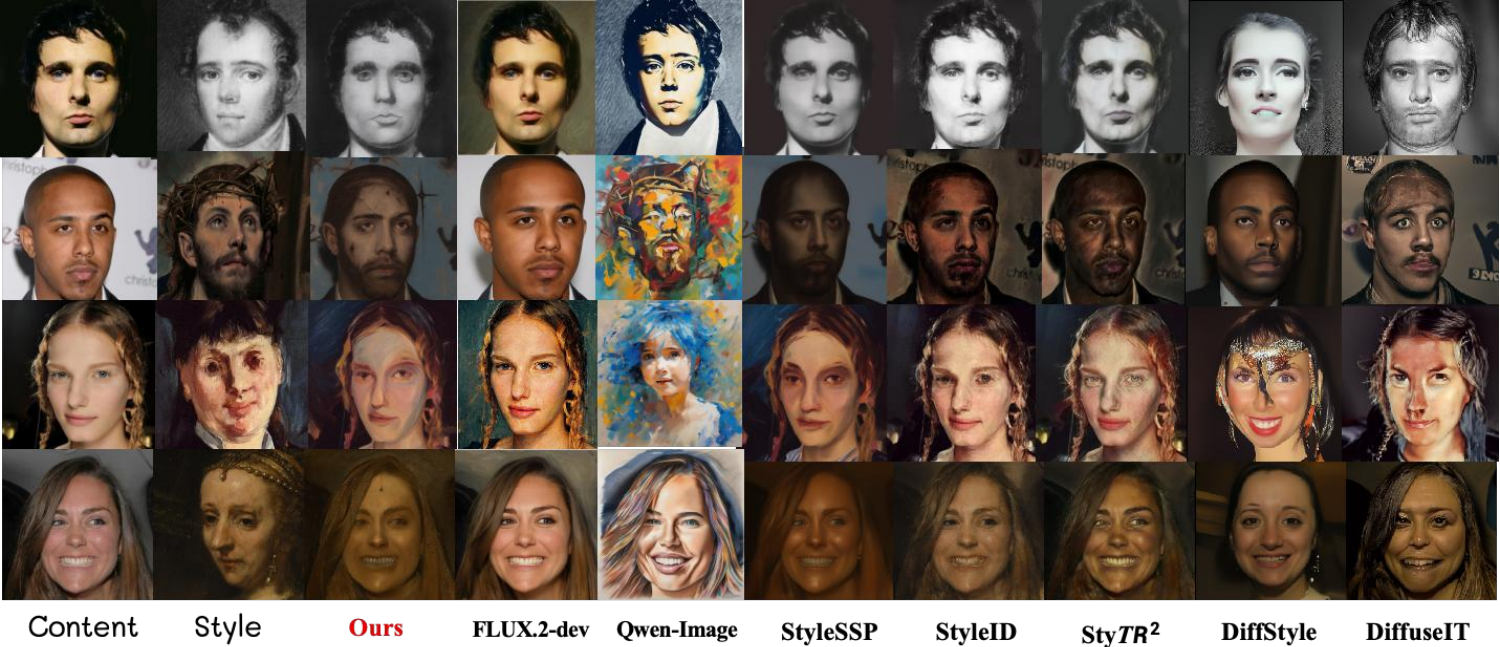}
    \caption{Visual comparison of style transfer results. Our \method better preserves facial identity while effectively transferring the artistic style compared to baseline methods.}
    \label{fig:main_style_transfer}
    \vspace{-16pt}
\end{figure*}
\textbf{Evaluation:}  We quantitatively evaluate our method along three complementary axes. Style alignment is measured with ArtFID~\cite{wright2022artfid}; content preservation is assessed using LPIPS~\cite{zhang2018unreasonable} and Content Loss (CS)~\cite{li2017universal}; and identity preservation is assessed by identity loss (ID), computed as one minus the dot product between normalized ArcFace~\cite{deng2019arcface} embeddings of the content and stylized images. %Thus, lower ID values indicate better identity preservation. 
For a comprehensive comparison, we benchmark \method against representative baselines drawn from three categories: (i) \emph{DiT-based editing models}, including Qwen-Image-Edit~\cite{wu2025qwen} and FLUX.2-dev~\cite{labs2025flux}; (ii) \emph{diffusion-based style transfer methods}, including StyleSSP~\cite{xu2025stylessp}, StyleID~\cite{chung2024style}, DiffuseIT~\cite{kwon2022diffusion}, InST~\cite{zhang2023inversion}, and DiffStyle~\cite{jeong2024training}; and (iii) \emph{conventional feed-forward methods}, including AesPA-Net~\cite{hong2023aespa}, CAST~\cite{zhang2022domain}, StyTR$^2$~\cite{deng2022stytr2}, and ArtFlow~\cite{an2021artflow}. %We quantitatively evaluate our method along three complementary axes. Style alignment is measured with ArtFID~\cite{wright2022artfid}; content preservation is assessed using LPIPS~\cite{zhang2018unreasonable} and Content Loss (CS)~\cite{li2017universal}; and identity preservation (ID) is computed as one minus the dot product between normalized ArcFace~\cite{deng2019arcface} embeddings of the content and stylized images. The baselines are drawn from three categories: (i) \emph{DiT-based editing models}, including Qwen-Image-Edit~\cite{wu2025qwen} and FLUX.1-dev~\cite{labs2025flux}; (ii) \emph{diffusion-based style transfer methods}, including StyleSSP~\cite{xu2025stylessp}, StyleID~\cite{chung2024style}, DiffuseIT~\cite{kwon2022diffusion}, InST~\cite{zhang2023inversion}, and DiffStyle~\cite{jeong2024training}; and (iii) \emph{conventional feed-forward methods}, including AesPA-Net~\cite{hong2023aespa}, CAST~\cite{zhang2022domain}, StyTR$^2$~\cite{deng2022stytr2}, and ArtFlow~\cite{an2021artflow}.
%We quantitatively evaluate our method using ArtFID~\cite{wright2022artfid} to measure style alignment, LPIPS~\cite{zhang2018unreasonable} and Content Loss~\cite{li2017universal} to assess content preservation, and an identity preservation metric (ID)~\cite{jeong2024training} based on ArcFace~\cite{deng2019arcface} embeddings.

%To quantitatively evaluate our \method, we employ a comprehensive set of metrics. For global performance evaluation, we adopt ArtFID~\cite{wright2022artfid}, which measures the distributional alignment between generated and reference images. For content preservation assessment, we utilize two complementary metrics: LPIPS~\cite{zhang2018unreasonable} and Content loss~\cite{li2017universal}, which measure perceptual and structural similarities between stylized outputs and their corresponding content images. Additionally, we employ an identity preservation metric (ID)~\cite{jeong2024training} that quantifies facial feature consistency through cosine similarity between ArcFace~\cite{deng2019arcface} embeddings of content and stylized images.
\begin{table*}[!h]
    \centering
    \small
    \vspace{-4pt}
    \caption{The style transfer results with using WikiArt and MetFace datasets as style reference.}
   \begin{threeparttable}
    \resizebox{0.95\textwidth}{!}{
    \begin{tabular}{c|cccc|cccc}
      % Method  & \multicolumn{8}{c|}{WikiArt} & \multicolumn{8}{c}{MetFace} \\
      \toprule
      ~ & \multicolumn{4}{c|}{\textbf{WikiArt}} & \multicolumn{4}{c}{\textbf{MetFace}} \\
       Method & ArtFID ($\downarrow$) & LPIPS ($\downarrow$) & ID ($\downarrow$) & CS ($\uparrow$) &  ArtFID ($\downarrow$) & LPIPS ($\downarrow$) & ID ($\downarrow$) & CS ($\uparrow$)\\
       \midrule
        AesPA-Net~\cite{hong2023aespa} & 28.3402 & 0.4076 & 0.2696 &  0.4254 & 29.1623 &  0.4853 &  0.2968 &  0.3948\\ %0.9926\\
        ArtFlow~\cite{an2021artflow} & \textbf{22.9778 } & 0.4933 & 0.2555  & 0.3215 & 30.2701 & 0.4717 & 0.2484 & 0.3932  \\ %0.9930  \\
        CAST~\cite{zhang2022domain} & 33.0048 & 0.3856 & 0.2832 & 0.4064 & 28.6567 & 0.4109 & 0.2877 & 0.4293 \\ %0.9926\\
        %EFDM & 22.7351 & 14.5203 & 0.46486 & 0.38939 & 0.49594 & 0.22399 & 0.00080 & 0.0395 &  0.3802\\
        StyTR$^2$~\cite{deng2022stytr2} & 24.2880 & 0.4068 & 0.1760 & 0.5181 &  26.6994 & 0.4104 & 0.1599 & 0.5503 \\ %0.9931 \\
    \midrule
    %\multicolumn{9}{c}{Diffusion-based Method} \\
    %\midrule
    InST~\cite{zhang2023inversion} & 29.8916 & 0.5975 & 0.8075 & 0.1862 & 33.4454 & 0.6084 & 0.8432 & 0.2447 \\%0.9903\\
    DiffStyle~\cite{jeong2024training} & 43.5315 & 0.6084 & 0.7988 & 0.3220 & 46.6953 & 0.6128 & 0.7824 & 0.3195\\%0.9931\\
    DiffuseIT~\cite{kwon2022diffusion} & 40.2349 & 0.5658 & 0.6619 & 0.2845 & 37.3332 & 0.5540 & 0.6542 & 0.2912 \\ %0.9913\\

    StyleID~\cite{chung2024style} & 24.0963 &  0.3785 & 0.1618 & 0.4493 & 26.6917 & 0.5283 & 0.1646 & 0.4283 \\%0.9926\\
    \midrule 
    FLUX.2-dev~\cite{labs2025flux} & 53.7948 & 0.7487 & 0.4146 & 0.2775 & 44.1780 & 0.6870 & 0.3766 &0.3056 \\
    Qwen-Image-Edit~\cite{wu2025qwen} &  35.4498 & 0.7196 & 0.4334 & 0.1335 & 33.1020 & 0.6946 & 0.3707 & 0.2008 \\
    \midrule
    \method (Ours) &  24.4906 & \textbf{0.2616} & \textbf{0.1237} & \textbf{0.5531} & \textbf{26.6203} & \textbf{0.3118} & \textbf{0.1424} & \textbf{0.5618}\\  
    \midrule
    \multicolumn{9}{c}{Diffusion-based methods with additional Control\tnote{1}} \\
    \midrule
    StyleSSP~\cite{xu2025stylessp} & 36.3980 & 0.4535 & 0.1707 & 0.4281& 35.6287 & \textbf{0.4079} & 0.1658&  \textbf{0.5322} \\
    \method$^*$ (Ours) & \textbf{35.1855} & \textbf{0.4523} & \textbf{0.1433} &\textbf{0.4572} & \textbf{31.2170} & 0.5228 &\textbf{0.1441}   & 0.4454\\
    \bottomrule 
    \end{tabular}
    }
    \begin{tablenotes}
    \item[1] Include both ControlNet~\cite{zhang2023adding} and IP adapter~\cite{ye2023ip-adapter}. 
    \end{tablenotes}
    \end{threeparttable}
    \label{tab:main_results}
    \vspace{-8pt}
\end{table*}
\textbf{Dataset:}
Our experiments utilize three datasets: CelebA~\cite{liu2015faceattributes} for real-world faces, MetFace~\cite{karras2020training} and WikiArt~\cite{tan2018improved} for artistic styles. Following standard protocol~\cite{chung2024style,deng2022stytr2}, we randomly sample 20 content images from CelebA and 40 style reference images from each artistic dataset. For WikiArt, we use images from WikiArt-Face~\footnote{https://huggingface.co/datasets/asahi417/wikiart-face} as the reference style facial images, reprocessing the images by cropping facial regions with 0.5 expansion ratio and resizing to 512$\times$512. And for MetFace, we just random select 40 images from MetFace~\footnote{https://github.com/NVlabs/metfaces-dataset} without any reprocessing.

%\textbf{Baselines:}
%To evaluate the effectiveness of our proposed \method, we compare against several state-of-the-art style transfer approaches. We select representative methods from two categories: (1) diffusion-based methods including StyleSSP~\cite{xu2025stylessp}, StyleID~\cite{chung2024style}, DiffuseIT~\cite{kwon2022diffusion}, InST~\cite{zhang2023inversion}, and DiffStyle~\cite{jeong2024training}; and (2) conventional methods including AesPA-Net~\cite{hong2023aespa}, CAST~\cite{zhang2022domain}, StyTR$^2$~\cite{deng2022stytr2}, and ArtFlow~\cite{an2021artflow}.

\subsection{Main Results}

As shown in Table~\ref{tab:main_results}, our \method demonstrates superior performance compared to existing CNN-based and diffusion-based approaches across multiple quantitative metrics. Specifically, on the WikiArt dataset, our approach achieves state-of-the-art results in LPIPS (0.2616), identity preservation (0.1237), and content similarity (0.5531). The performance advantages are further validated on the MetFace dataset, where our \method exhibits leading scores across all evaluation metrics, including ArtFID, LPIPS, and identity preservation. 

%While StyleID and StyTR$^2$ show competitive identity preservation, they exhibit suboptimal style transfer capabilities. Other diffusion-based methods yield higher ArtFID and LPIPS scores, indicating limitations in both stylization and content preservation. 
Our third-place WikiArt ArtFID indicates that ArtFlow and StyleID stylize more strongly, at the cost of larger identity and content changes; \method thus offers the best identity and content preservation at competitive stylization strength, rather than dominating every metric. Visual comparisons (Figure~\ref{fig:main_style_transfer}) further validate our method's superior performance in balancing identity preservation and style transfer. Visual comparisons (Figure~\ref{fig:main_style_transfer}) further validate our method's superior performance in balancing identity preservation and style transfer. More visualization results can refer to Appendix~\ref{app:visualization}.

\begin{figure*}[!h]
\vspace{-4pt}
    \centering
    \includegraphics[width=\linewidth]{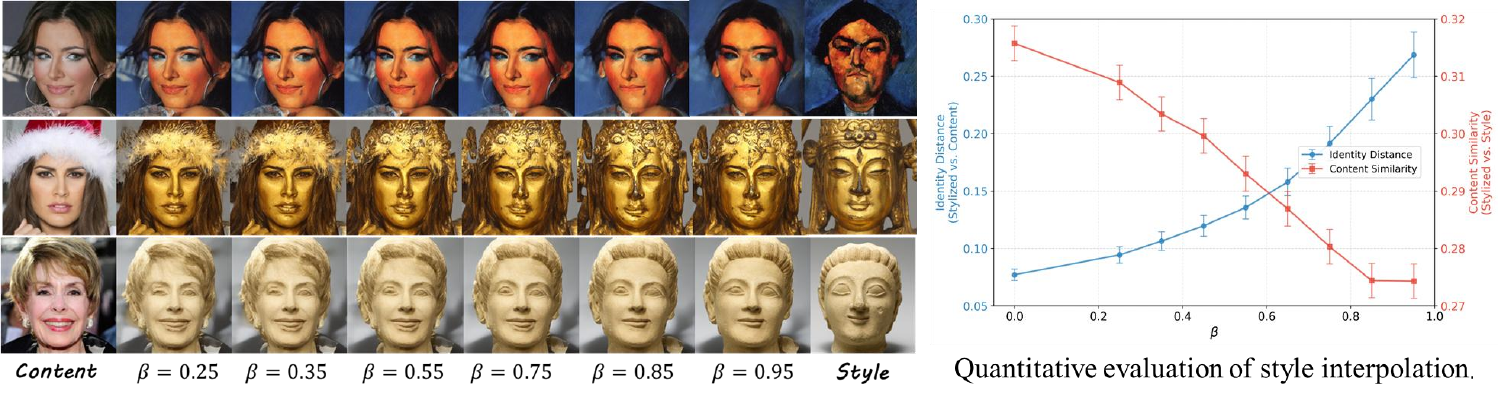}
    \caption{Style interpolation results showing smooth transitions between different artistic styles.}
    \label{fig:interpolation}
    \vspace{-16pt}
\end{figure*}
\subsection{Ablation Study}
\begin{figure}[h]
\vspace{-8pt}
  \begin{minipage}[c]{0.4\textwidth}
    \centering
    \captionof{table}{Ablation study results comparing our full method against variants with key components removed. Experiments are evaluated with using the MetFace dataset as style images.}
     \resizebox{\linewidth}{!}{
    \begin{tabular}{c|cccc}
    \toprule
    Method &  ArtFID ($\downarrow$) & LPIPS ($\downarrow$) & ID ($\downarrow$) & CS ($\uparrow$) \\
    \midrule
      w/o Sty. Interp. Init.  &  26.4807 & 0.3647 & \textbf{0.1250} & 0.4933  \\
      w/o Sty. Inject Ctrl & \textbf{26.3421} & 0.3861  & 0.1470 &  0.4735\\
      w/o both & 26.7091 & 0.3877 & 0.1473 & 0.4715\\
    \midrule
    \method &  26.6203 & \textbf{0.3118} & \text{0.1424} & \textbf{0.5618} \\
    \bottomrule
    \end{tabular}
    }
     \label{tab:ablation}
  \end{minipage}
  \hfill
  \begin{minipage}[c]{0.58\textwidth}
    \centering
    \includegraphics[width=\linewidth]{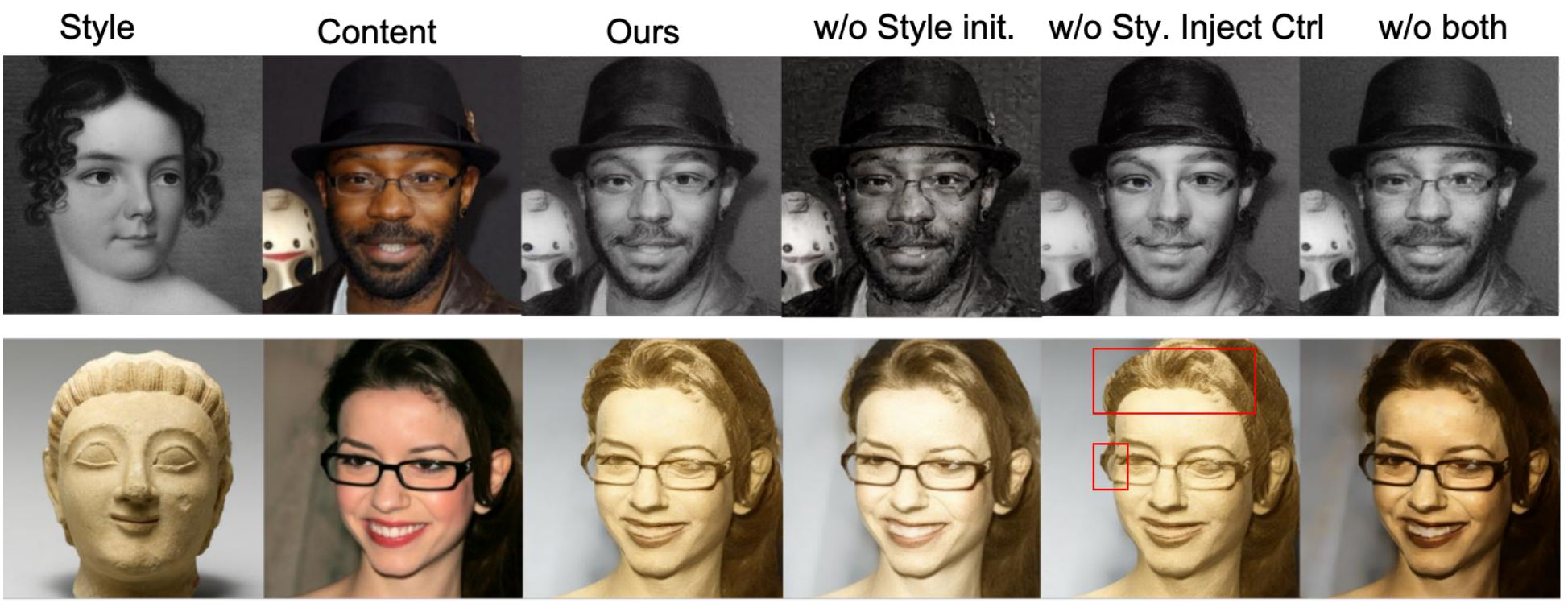}
    \caption{The visualization of ablation study.}
    \label{fig:ablation}
  \end{minipage}
   \vspace{-8pt}
\end{figure}

We perform ablation studies to understand the contribution of each component of our \method. We remove the style interpolation initial (see Section~\ref{sec:style_interp}) and the style injection control (see Section~\ref{sec:style_injection_control}) to see the performance of our \method. The results are shown in Table~\ref{tab:ablation}.
The ablation study results in Table~\ref{tab:ablation} analyze our \method's key components: style interpolation initialization and style injection control. Removing style interpolation initialization maintains comparable ArtFID and improves ID from 0.1424 to 0.1250, but worsens LPIPS and CS, revealing a trade-off rather than uniform degradation.
%The ablation study results in Table~\ref{tab:ablation} analyze our \method's key components: style interpolation initialization and style injection control. Removing style interpolation initialization maintains comparable ArtFID but degrades perceptual quality and identity preservation.
%Figure~\ref{fig:ablation} shows this degradation visually, particularly in facial features where stylized outputs lack coherent adaptation of eye shapes and skin textures compared to our full \method.
Analysis of the style injection control ablation reveals that its removal, while marginally improving ArtFID, significantly degrades other performance metrics, indicating compromised content preservation. The ablation of both components yields the most severe degradation, empirically validating their synergistic relationship.

\subsection{Linear Style Interpolation Analysis}
%As introduced in Section~\ref{sec:style_injection_control}, \method supports continuous style injection via a scalar parameter $\beta \in [0,1]$. Figure~\ref{fig:interpolation} shows that varying $\beta$ induces smooth, nearly linear transitions from content to style: small values ($\beta=0.2, 0.4$) inject subtle stylistic cues while preserving identity, whereas larger values ($\beta=0.6, 0.8$) amplify artistic characteristics without disrupting facial structure. This coherence stems from the geodesic-based interpolation initialized with $\delta=0.9$, which fuses content and style along the latent manifold rather than in Euclidean space, and is quantitatively corroborated by the monotonic progression of identity loss across $\beta$. Together, these results demonstrate that \method affords precise and continuous control over the content-style trade-off while maintaining semantic consistency.
As introduced in Section~\ref{sec:style_injection_control}, \method supports continuous style injection via a scalar parameter $\beta \in [0,1]$. Figure~\ref{fig:interpolation} shows that varying $\beta$ induces smooth, nearly linear transitions from content to style: small values ($\beta=0.2, 0.4$) inject subtle stylistic cues while preserving identity, whereas larger values ($\beta=0.6, 0.8$) amplify artistic characteristics without disrupting facial structure. This coherence stems from the $\delta$-initialized interpolation with energy-based refinement, and is quantitatively corroborated by the monotonic progression of identity loss across $\beta$. Together, these results demonstrate that \method affords precise and continuous control over the content-style trade-off while maintaining semantic consistency.

\begin{figure*}[!h]
    \centering
    \vspace{-12pt}
    \includegraphics[width=\linewidth]{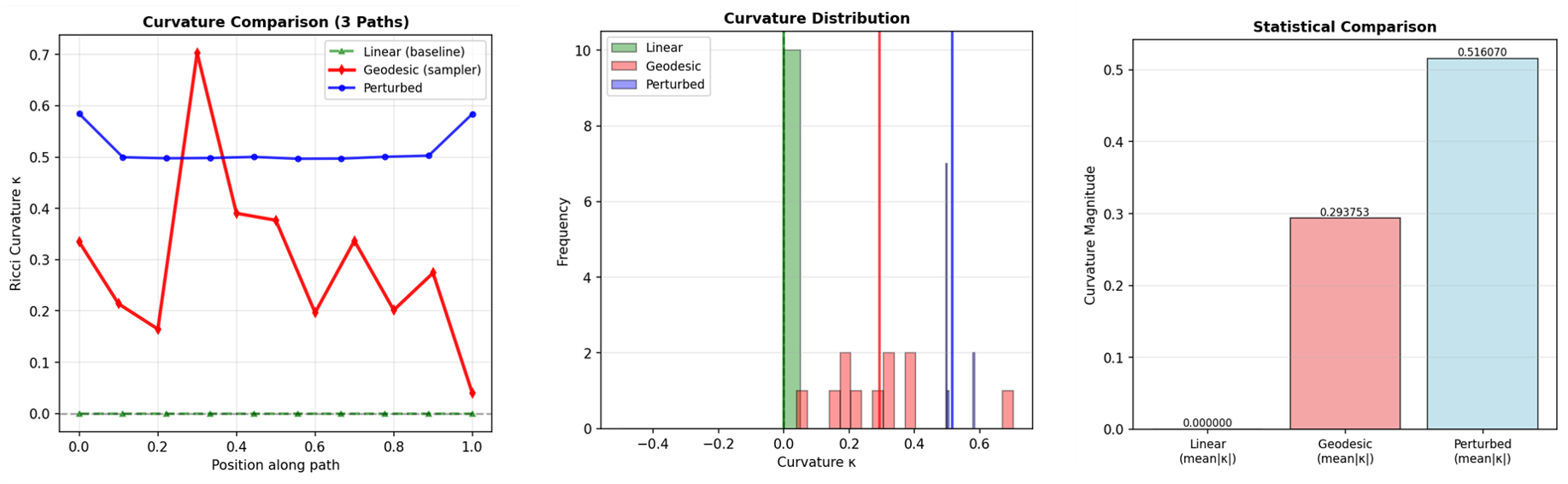}
    \caption{Discrete path-curvature statistics for linear, energy-refined, and randomly perturbed interpolation trajectories.}
    \label{fig:local_isometry}
    \vspace{-16pt}
\end{figure*}

\subsection{Geometric Properties of Interpolation Paths}\label{sec:local_isometry}
To examine whether the refinement produces a smooth and structured departure from linear interpolation, we compare three trajectories between content and style latents: (i) linear interpolation $z(t)=(1-t)z_c+t z_s$, (ii) the energy-refined trajectory formed by optimizing points at the same interpolation ratios, and (iii) a randomly perturbed trajectory. For consecutive sampled points, we compute a finite-difference path-curvature proxy from the change in local trajectory direction. This is a diagnostic of trajectory shape, not an estimator of the manifold's intrinsic Ricci curvature.

%\rev{To validate that geodesic paths provide superior interpolation compared to simple linear paths, we analyze the geometric properties of different interpolation strategies in the diffusion latent space. Our theoretical framework relies on the local isometry property, which requires well-behaved manifold geometry. We empirically verify this through curvature analysis and comparison with alternative interpolation schemes. We compute Ricci curvature along two types of paths: (1) linear interpolation $z(t) = (1-t)z_c + t z_s$ in latent space,  (2) geodesic paths computed via our energy minimization framework, and (3) randomly perturbed paths.}
As shown in Fig.~\ref{fig:local_isometry}, the linear baseline has near-zero path curvature by construction, while random perturbations have the largest and most variable curvature. The energy-refined trajectory lies between these extremes: it departs from the straight line but remains substantially smoother than the perturbed trajectory. This provides empirical support that the optimization introduces a structured local correction rather than an unstable deformation; it does not by itself prove local isometry or bounded intrinsic curvature.

\subsection{Method Hyperparameter Analysis}
\begin{figure}[h]
\begin{minipage}[b]{0.48\linewidth}
\vspace{-12pt}
    \centering
    \includegraphics[width=\linewidth]{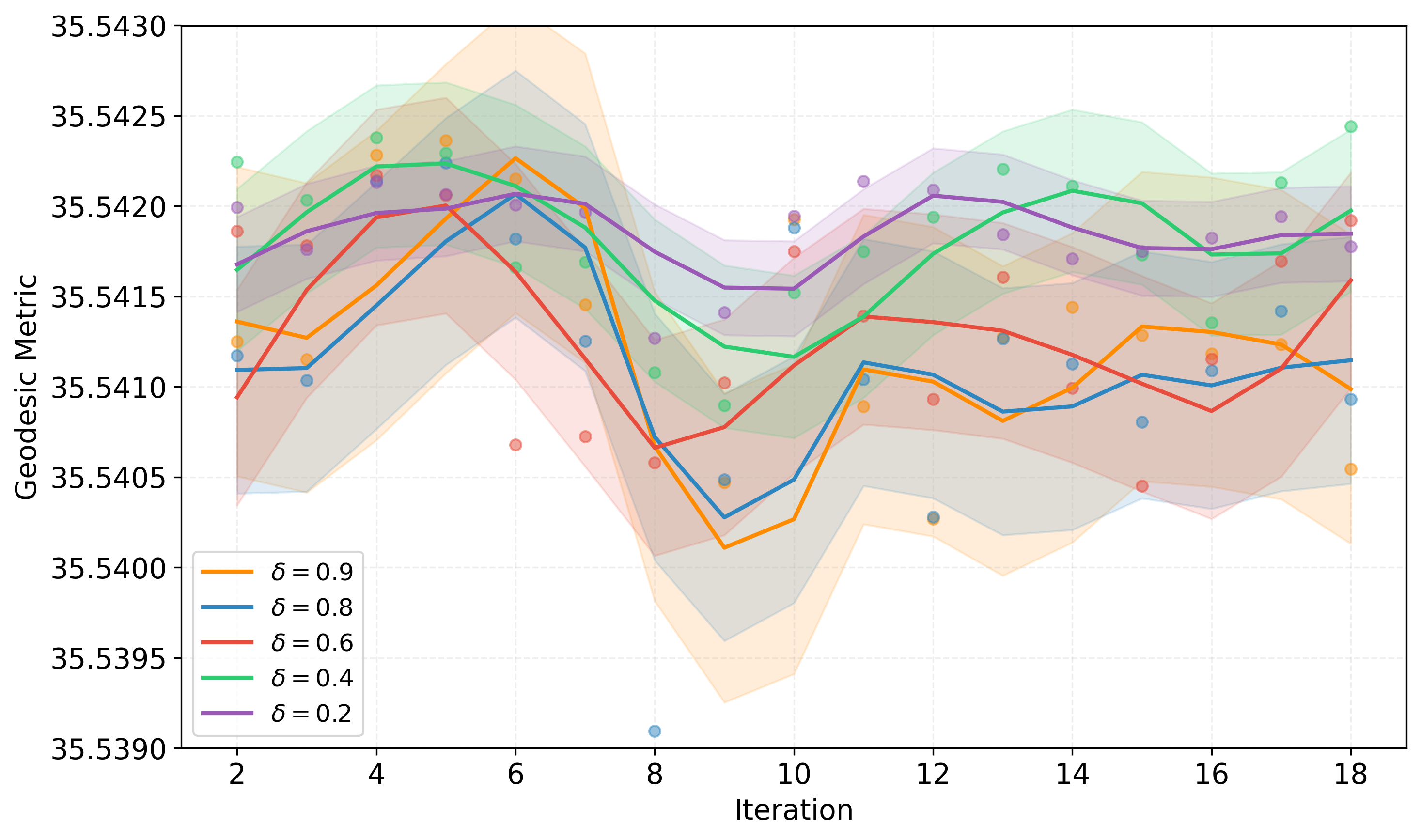}
    \caption{The effect of $\delta$. 
    %on identity preservation and content similarity.
    }
    \label{fig:delta_effect}
\end{minipage}
\begin{minipage}[b]{0.48\linewidth}
    \centering
    \includegraphics[width=\linewidth]{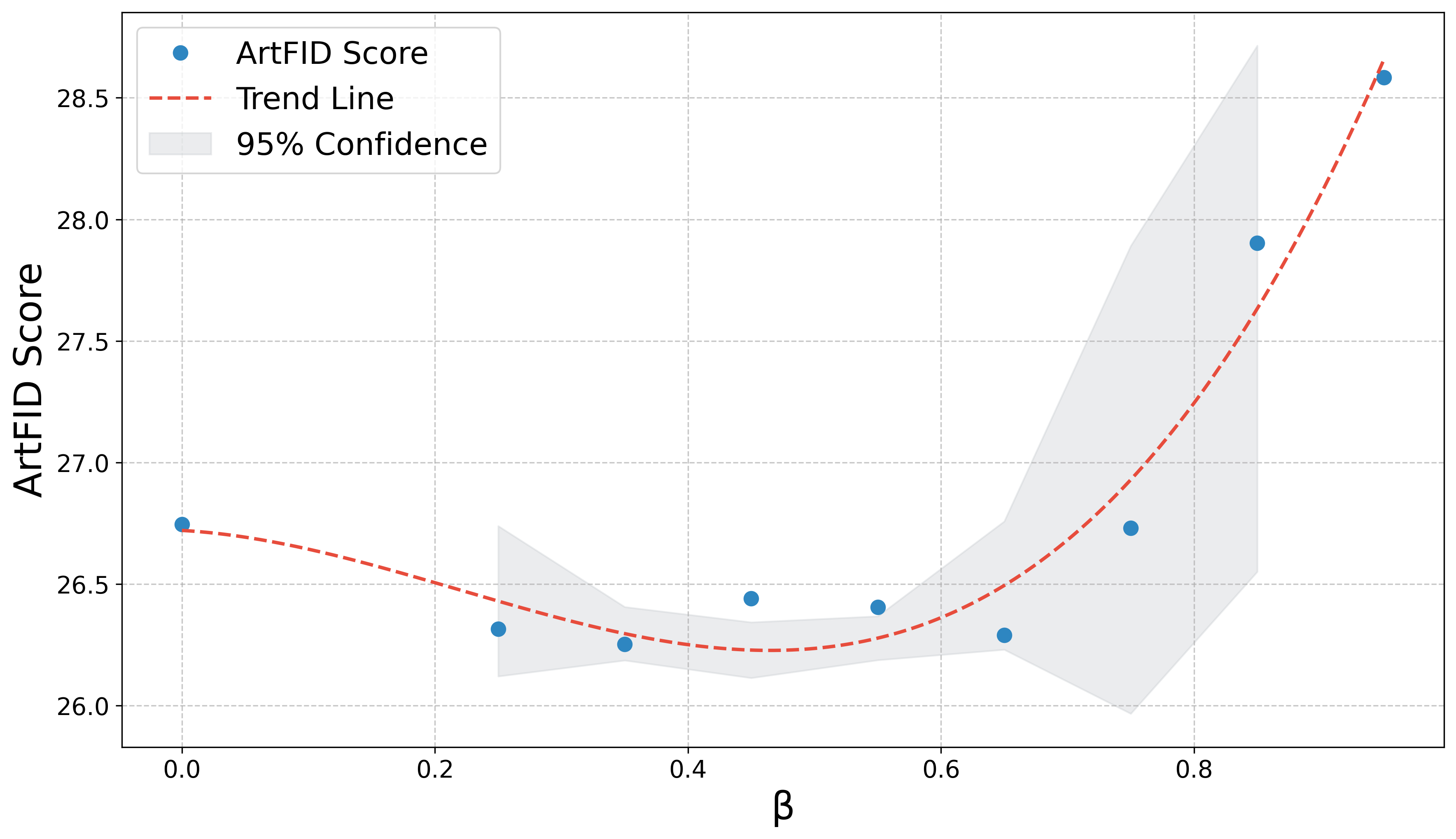}
    \caption{The effect of $\beta$. 
    %on style transfer performance.
    }% the ArtFID value changes are plotted along the increase of $\beta$.}
    \label{fig:beta_effect}
\end{minipage}
\vspace{-16pt}
\end{figure}

We analyze the influence of the interpolation parameters $\delta$ and $\beta$ on style transfer performance. As shown in Figure~\ref{fig:delta_effect}, $\delta$ controls the stability of the geodesic path optimization. Larger values lead to more stable geodesic trajectories, while smaller values introduce instability. We empirically set $\delta=0.9$ to ensure stability without excessive style injection. The parameter $\beta$ balances style strength and content preservation, as illustrated in Figure~\ref{fig:beta_effect}. ArtFID scores initially improve with increasing $\beta$ but then degrade as the stylized image diverges from the original content. We find $\beta=0.25$ provides the best trade-off between style transfer quality and content fidelity. The step size $\lambda$ and iteration budget are kept small so that the refinement remains a local correction (Section~\ref{sec:style_interp}).
\section{Conclusion}
We introduced \method, a training-free framework for controllable facial stylization. The main idea is simple: linear interpolation ignores the curved geometry learned by a diffusion model. Starting from a ratio-selected linear interpolant, \method therefore makes a small, geodesic-guided correction by minimizing a two-segment energy in the model's hidden representation. The pullback metric provides a local geometric justification for this correction; our method does not claim to recover a globally optimal geodesic.
We apply the same refinement at two complementary levels. Latent refinement controls the overall content--style blend, while self-attention query refinement injects finer stylistic details during denoising. Together, they provide continuous control over stylization strength without additional training. Experiments on WikiArt and MetFace show strong identity and content preservation at competitive style alignment, and the ablations confirm that the two components are complementary. These results suggest that local geometric refinement is a practical way to improve controllable stylization and may extend to broader visual domains and diffusion architectures.

\bibliography{main}
\bibliographystyle{iclr2027_conference}

\newpage
\clearpage
\setcounter{page}{1}
\appendix
\begin{figure}[h]
    \centering
      \vspace{-16pt}
    \includegraphics[width=\linewidth]{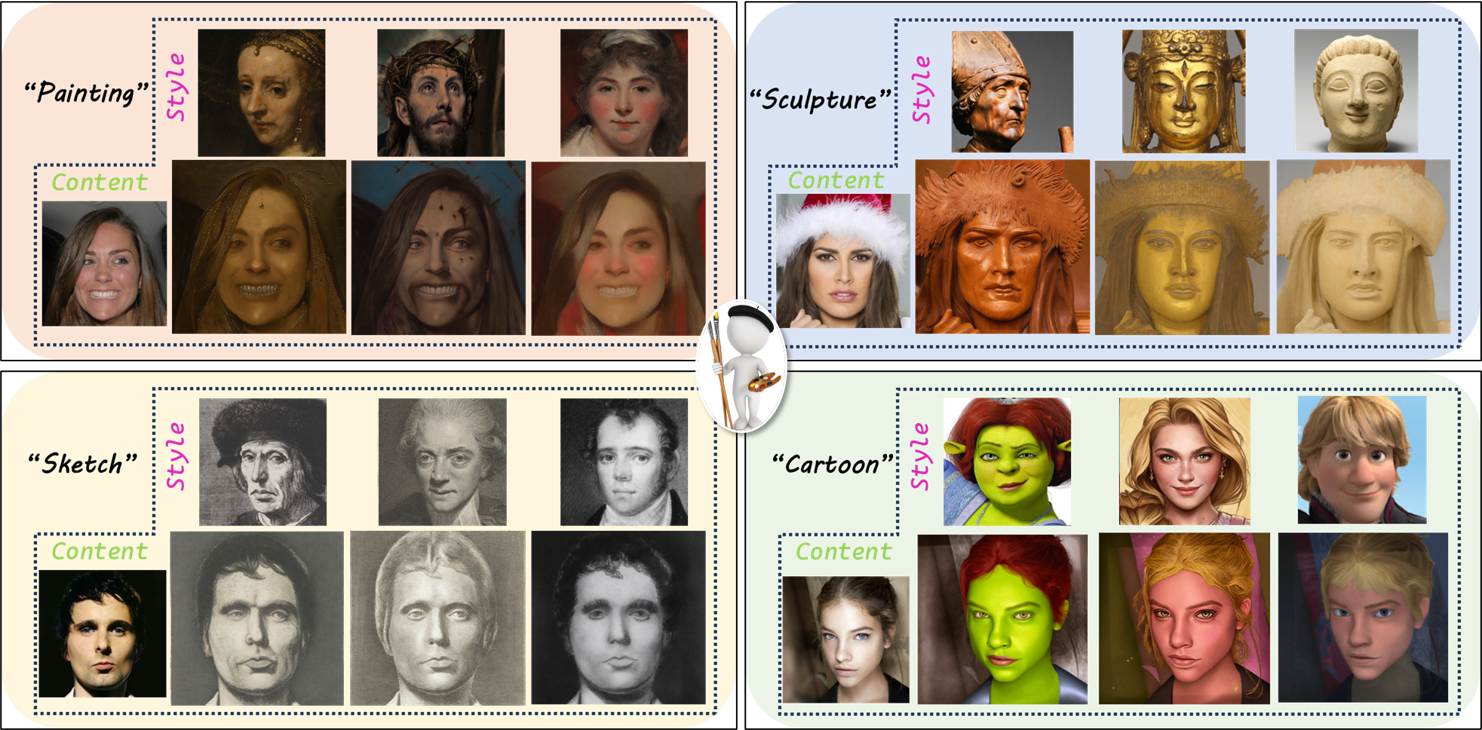}
    %\caption{Caption}
    \label{fig:placeholder}
    \vspace{-16pt}
\end{figure}
%\maketitlesupplementary
%\section{Source Code}
%\section{Broader Impact}
%This work focuses on controllable stylization of human facial images, which has potential impact in creative media, digital art, .etc. As facial manipulation technologies may also be misused for identity impersonation or deceptive content generation, ethical considerations are important. Our method is training-free and does not extend the capabilities of existing generative models, but rather provides a principled control mechanism over their behavior. We encourage responsible use and hope that the proposed geometric perspective contributes to more transparent and interpretable generative modeling.
\section{Limitations}
Our work opens up several promising directions for future research. First, the principles of our dynamic style injection control could be extended to other domains beyond facial images. Second, the linear interpolation mechanism could be further enhanced to handle more complex style transformations. Finally, investigating the integration of our approach with text-guided diffusion models could enable more intuitive control over the stylization process.

\section{Geodesic Path Computation}~\label{sec:geodesic}
%\subsection{Geodesic Path Computation}
%The latent space of a generative model $g$ can be interpreted as a Riemannian manifold~\cite{shao2018riemannian,park2023understanding}. When two points from the coordinate space $\mathcal{X}$ are mapped into this latent space by the generative model, there exists a unique geodesic path connecting them on the Riemannian manifold. Given two points $x_1$ and $x_2$ in the coordinate space, we define a path $\gamma: [0,1] \rightarrow \mathcal{X}$ connecting them, where $\gamma(0) = x_1$ and $\gamma(1) = x_2$. This path corresponds to a unique curve on the Riemannian manifold $\mathcal{M}$, given by $g \circ \gamma(t) \in \mathcal{M}$. The curve $\gamma$ follows a direction from $x_1$ to $x_2$, and this directional path is preserved when mapped to the corresponding curve on the Riemannian manifold $\mathcal{M}$ through the generative model $g$.
This section provides general background on geodesics under the pullback metric. As stated in Section~\ref{sec:style_interp}, our method does not solve the geodesic equation derived here; it only uses the discrete two-chord energy derived at the end of this section. Where the Jacobian of $g$ has full column rank, the latent space can be endowed with a Riemannian metric pulled back from the hidden-state space~\cite{shao2018riemannian}. Here, as in the main text, $g(x)=g_d(x,t,c)$ is the frozen U-Net down-path/bottleneck hidden-state map at fixed timestep and conditioning, and $\mathcal{M}=g(\mathcal{X})$ is its image. Within a neighborhood where this holds, nearby points are connected by locally length-minimizing curves (geodesics); we do not assume that such a geodesic exists or is unique globally. Consider two points $x_1, x_2 \in \mathcal{X}$ in the coordinate space. Let $\gamma: [0,1] \rightarrow \mathcal{X}$ be a smooth path such that $\gamma(0) = x_1$ and $\gamma(1) = x_2$. The composition map $g \circ \gamma: [0,1] \rightarrow \mathcal{M}$ induces a curve on $\mathcal{M}$, where $g \circ \gamma(t) \in \mathcal{M}$ for all $t \in [0,1]$. On a neighborhood where $g$ is a diffeomorphism onto its image (Section~\ref{sec4.1}), this correspondence between paths in $\mathcal{X}$ and curves on $\mathcal{M}$ is one-to-one. To characterize the geometric relationship between the latent space and the manifold, we have the following definition.
% To characterize the geometry of this manifold, we define a Riemannian metric tensor as a smooth, symmetric, and positive definite matrix field $G(x)$ that acts on the tangent space at each point of the latent coordinate space:
% \[
% G(x) = J_{g}(x)^{\top}J_{g}(x),
% \]
% where $g(x) \in \mathcal{M}$ denotes the mapping from coordinate space to Riemannian manifold, and $J_{g}(x)$ represents the Jacobian matrix of $g$ at point $x$.

\begin{definition}[Pullback metric $\mathsf g$ on $\mathcal{X}$ induced by the generator]
Let $g:\mathcal{X}=\mathbb{R}^n\to\mathcal{M}\subset\mathbb{R}^p$ be a continuously differentiable map with Jacobian $J_g(x)\in\mathbb{R}^{p\times n}$.
For each $x,u,v\in \mathcal{X}$, define
\[
  \mathsf g_x(u,v)\ :=\ \big\langle J_g(x)\,u,\; J_g(x)\,v\big\rangle_{\mathbb{R}^p}.
\]
In local coordinates, the matrix representation of $\mathsf g_x$ is
\[
  G(x)\ =\ J_g(x)^{\top}J_g(x),
\]
so that for $u,v\in\mathbb{R}^n$,
\[
  \mathsf g_x(u,v)\ =\ u^{\top}G(x)\,v,\qquad
  \|u\|_{g,x}^2\ =\ u^{\top}G(x)\,u\ =\big\|J_g(x)\,u\big\|_2^2.
\]
If $J_g(x)$ has full column rank, then $\mathsf g_x$ is positive definite, and $\mathsf g$ is a Riemannian metric on any open set where this holds.
\end{definition}

Given this metric, we can formally define the arc length of a smooth path $\gamma$ through the Riemannian line integral:
\begin{equation}
\small
\begin{aligned}
    L(\gamma) = \int_{0}^{1} \sqrt{{\frac{d\gamma(t)}{dt}}^{\top} G(\gamma (t)) \frac{d{\gamma}(t)}{dt}} dt = \\ \int_{0}^{1}\sqrt{\mathsf  g_{\gamma(t)}\Big(\frac{d\gamma(t)}{dt},\frac{d\gamma(t)}{dt}\Big)} dt.
\end{aligned}
\end{equation}
This integral measures the length of the path under the pullback metric. A geodesic minimizes this arc length locally; for constant-speed parametrizations this is equivalent to minimizing the energy function, which accumulates the squared speed:
\begin{equation}
    E(\gamma) = \frac{1}{2}\int_{0}^{1} {\frac{d\gamma(t)}{dt}}^{\top} G(\gamma (t))\frac{d{\gamma}(t)}{dt} dt.
    \label{eq:energy}
\end{equation}
Using energy is standard because it is smooth and makes the calculus of variations straightforward. Then we express the Christoffel symbols of the metric $G$, denoted as $\Gamma^i_{jk}$:
\begin{equation}
    \Gamma^i_{jk} = \frac{1}{2}G_{il}^{-1}(\frac{\partial G_{lj}}{\partial x^k} + \frac{\partial G_{lk}}{\partial x^j} - \frac{\partial G_{jk}}{\partial x^l}),
\end{equation}
where $i,j,k$ are indices of the coordinate space. Operationally, they tell us how to differentiate tangent vectors along the manifold. By taking a variation of the energy function in Equation~\ref{eq:energy} and applying the calculus of variations, we derive the Euler-Lagrange equation:
\begin{equation}
    \frac{d^{2}\gamma^i}{dt^2} = -\Gamma^i_{jk} \frac{d{\gamma^j}}{dt}\frac{d{\gamma}^k}{dt}
    \label{eq:vari_eng}.
\end{equation}
This second-order ordinary differential equation is the stationary condition of the energy functional; similar functional optimization also arises in optimal transport and in distributionally robust optimization~\citep{chung2025unbalanced,li2026extreme}. In principle, a geodesic with fixed endpoints can be computed by solving this boundary-value problem numerically. This requires the Christoffel symbols, i.e., derivatives of $G = J_g^{\top}J_g$, which is intractable for a diffusion U-Net. Our method therefore does not solve Equation~\ref{eq:vari_eng}; instead, it optimizes a single intermediate point under the discrete energy below. 

Then we derive the approximation to $E$ in Equation (\ref{eq5}). Let $y(t):=g(\gamma(t))\in\mathcal M$ and recall the continuous energy on $(\mathcal X,\mathsf g)$ equals the kinetic energy of $y$ on $\mathcal M$:
\[
E(\gamma)=\frac12\int_0^1 \big\|\dot y(t)\big\|_2^2\,dt
\quad\text{(a.e.\ in $t$, since $g\in C^1$, $\gamma\in AC$).}
\]
Fix a split time $\tau\in(0,1)$ and split $[0,1]$ into two subintervals $[0,\tau]$ and $[\tau,1]$.
Write $y_c:=g(x_c)=y(0)$, $y_m:=g(x_m)=y(\tau)$, $y_s:=g(x_s)=y(1)$. Under the initialization $x_m=\delta x_c+(1-\delta)x_s$ used in the main text, the initial point lies at $\tau = 1-\delta$. Then
\[
E(\gamma)=\frac12\!\left(\int_{0}^{\tau}\!\!\|\dot y(t)\|_2^2\,dt
+\int_{\tau}^{1}\!\!\|\dot y(t)\|_2^2\,dt\right).
\]

Assume on each subinterval the mapped curve $y(t)$ is well approximated by a straight segment with constant speed. Using forward/backward finite differences,
\[
\dot y(t)\approx
\begin{cases}
\dfrac{y_m-y_c}{\tau}, & t\in[0,\tau],\\[6pt]
\dfrac{y_s-y_m}{1-\tau}, & t\in[\tau,1].
\end{cases}
\]
Plugging into the energy gives the standard two-segment discretization:
\begin{align*}
E(\gamma)
&\approx \frac12\left(
\tau\Big\|\frac{y_m-y_c}{\tau}\Big\|_2^2
+(1-\tau)\Big\|\frac{y_s-y_m}{1-\tau}\Big\|_2^2\right)\\
&=\frac12\left(\frac{1}{\tau}\|y_m-y_c\|_2^2+\frac{1}{1-\tau}\|y_s-y_m\|_2^2\right). \tag{A.1}
\end{align*}

Bringing (A.1) over the common denominator $2\tau(1-\tau)$ yields
\[
E(\gamma)
=\frac{1}{2\tau(1-\tau)}\Big((1-\tau)\|y_m-y_c\|_2^2+\tau\|y_s-y_m\|_2^2\Big). \tag{A.2}
\]
Defining
$\alpha_\tau \;:=\; \frac{1}{2\tau(1-\tau)}$,
we obtain the compact form
\[
E(\gamma)\ \approx\ \alpha_\tau\Big((1-\tau)\,\|g(x_m)-g(x_c)\|_2^2+\tau\,\|g(x_s)-g(x_m)\|_2^2\Big). \tag{A.3}
\]
For an equal time split ($\tau=\tfrac12$), (A.3) reduces to
\[
E(\gamma)\ \approx\ \|g(x_m)-g(x_c)\|_2^2+\|g(x_s)-g(x_m)\|_2^2,
\]
i.e., the unweighted two-chord sum (here $\alpha_\tau=1$).

For the implementation in the main paper, $x_m=\delta x_c+(1-\delta)x_s$, so its location along the content-to-style segment is $\tau=1-\delta$. Substituting this relation into (A.3) gives
\[
E(\gamma)\approx\alpha\Big(\delta\|g(x_m)-g(x_c)\|_2^2
+(1-\delta)\|g(x_s)-g(x_m)\|_2^2\Big),
\]
where the common scale is absorbed into $\alpha$. The unconstrained mapped-space minimizer is $g(x_m)=\delta g(x_c)+(1-\delta)g(x_s)$. If this target lies outside $g(\mathcal X)$, optimization over $x_m$ instead seeks a local projection of the target onto $g(\mathcal X)$.

\section{Rank-One vs.\ Top-$k$ Singular Directions} \label{app:topk}
\section{Rank-One vs.\ Top-$k$ Singular Directions} \label{app:topk}
The update in Equation~\ref{eq:rank1_update} retains the component of the pullback gradient along the leading right singular direction $v_1$ of $J_g$ and discards the remaining directions, which could in principle carry subtle structural information relevant to fine facial details. To measure what is lost, we replace the rank-one update by its top-$k$ counterpart, $\eta_m^{(k)} = -2\alpha \sum_{i=1}^{k} v_i\,(J_g v_i)^{\top} r_m$, where $\{v_i\}_{i=1}^{k}$ are obtained by block power iteration on $J_g^{\top}J_g$ with orthonormalization, and compare $k\in\{1,2,4\}$ with all other settings unchanged.

\begin{table}[h]
    \centering
    \caption{Rank-one vs.\ top-$k$ singular directions (MetFace style references, SD v1.4). $k=1$ is the default setting of \method, identical to Table~\ref{tab:main_results}.}
    \label{tab:topk}
    \begin{tabular}{ccccc}
    \toprule
    $k$ & ArtFID ($\downarrow$) & LPIPS ($\downarrow$) & ID ($\downarrow$) & Time (s/image) \\
    \midrule
    1 (default) & 26.6203 & 0.3118 & 0.1424 & \textbf{9.71} \\
    2 & 26.51 & 0.3103 & 0.1418 & 15.4 \\
    4 & \textbf{26.49} & \textbf{0.3099} & \textbf{0.1416} & 25.8 \\
    \bottomrule
    \end{tabular}
\end{table}

As shown in Table~\ref{tab:topk}, increasing $k$ from 1 to 2 and 4 changes ArtFID, LPIPS, and ID only marginally, while the runtime grows almost linearly with $k$. In this evaluated regime, the task-relevant update is adequately captured by $v_1$, and the rank-one choice is a reasonable quality--efficiency trade-off. We note that this conclusion is specific to this regime and to the metrics considered; in strongly curved regions or for small spectral gaps, additional singular directions may matter.

\section{Background}
\subsection{Attention-based Style Injection}

Attention-based style injection~\cite{chung2024style} is a method used to incorporate style information into the latent representation of content images, offering a flexible and precise control over style transfer. This approach combines DDIM inversion with Self-Attention (SA) to effectively transfer style while maintaining content structure.

In this framework, the latents for both the content and style images are first obtained through DDIM inversion. Given a predefined set of timesteps \( t \in \{0, \dots, T\} \), the content latent \( \mathbf{z}_c^0 \) and style latent \( \mathbf{z}_s^0 \) undergo a deterministic forward process, progressively transforming into Gaussian noise distributions at timestep \( t = T \). During this inversion process, we extract and store the query features \( Q_t^c \) from the content image and the key-value pairs \( (K_t^s, V_t^s) \) from the style image at each timestep, which encode the structural and stylistic information respectively.

The stylization process begins by initializing the stylized latent noise \( \mathbf{z}_{cs}^T \) with the content latent noise \( \mathbf{z}_c^T \). Style transfer is then achieved through attention feature modulation, where the collected style features \( (K_t^s, V_t^s) \) are injected into the Self-Attention layer, replacing the original content features \( (K_t^{cs}, V_t^{cs}) \). This feature substitution enables precise style integration while preserving the content's structural integrity in the latent representation.

The synergy between DDIM inversion and attention-based feature modulation provides a theoretically grounded approach for style transfer, offering fine-grained control over the stylization process while maintaining the mathematical properties of both the diffusion and attention mechanisms.

\subsection{DDIM Inversion}

DDIM inversion~\cite{} is a forward process that progressively adds estimated noise to an image from timestep 0 to T, enabling the reconstruction of the noise schedule that would generate the image. Unlike the sampling process which denoises from T to 0, DDIM inversion follows the forward path to obtain the specific noise that characterizes the input image.

Given a clean image \(\mathbf{x}_0\), DDIM inversion follows a deterministic forward process to progressively add noise using the learned noise predictor \(\epsilon_\theta\). For timestep \(t\), the forward update rule is defined as:
\[
\mathbf{x}_{t+1} = \sqrt{\bar{\alpha}_{t+1}}\mathbf{x}_0 + \sqrt{1 - \bar{\alpha}_{t+1}}\epsilon_t
\]
where \(\bar{\alpha}_t = \prod_{s=1}^t \alpha_s\) represents the cumulative product of noise scheduling coefficients, and \(\epsilon_t\) is obtained through the noise predictor \(\epsilon_\theta(\mathbf{x}_t, t)\). This process iteratively continues from \(t=0\) to \(T\), generating a sequence of increasingly noisy latents \(\{\mathbf{x}_t\}_{t=1}^T\) that encode the image's characteristics in the model's latent space. The resulting noise schedule and latents can then be used for controlled generation through the reverse process.

\begin{figure}[h]
    \centering
    \includegraphics[width=\linewidth]{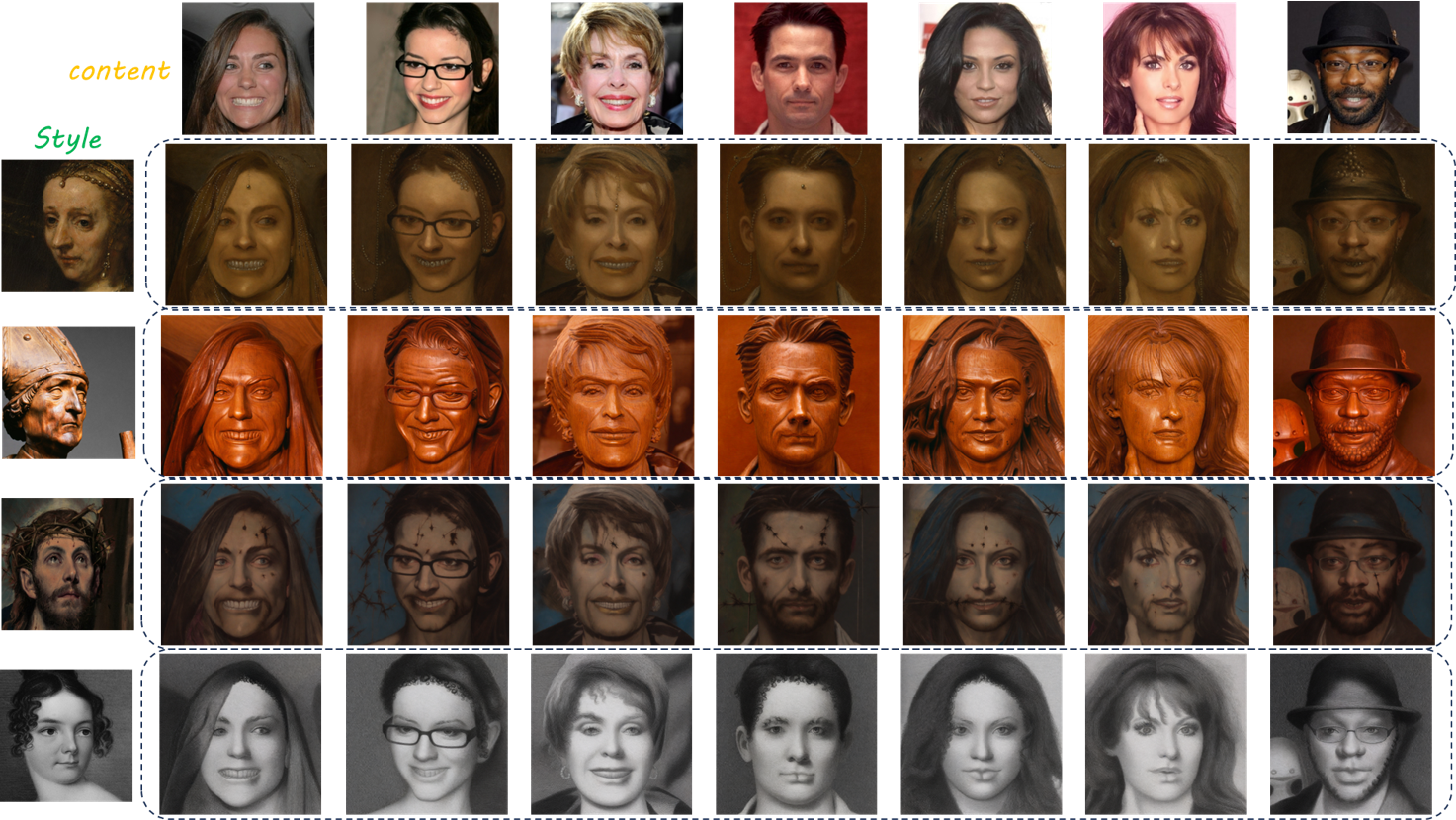}
    \caption{More Visualization Samples of the proposed \method$^*$.}
    \label{fig:add_vis_control}
\end{figure}

\begin{figure}[h]
    \centering
    \includegraphics[width=\linewidth]{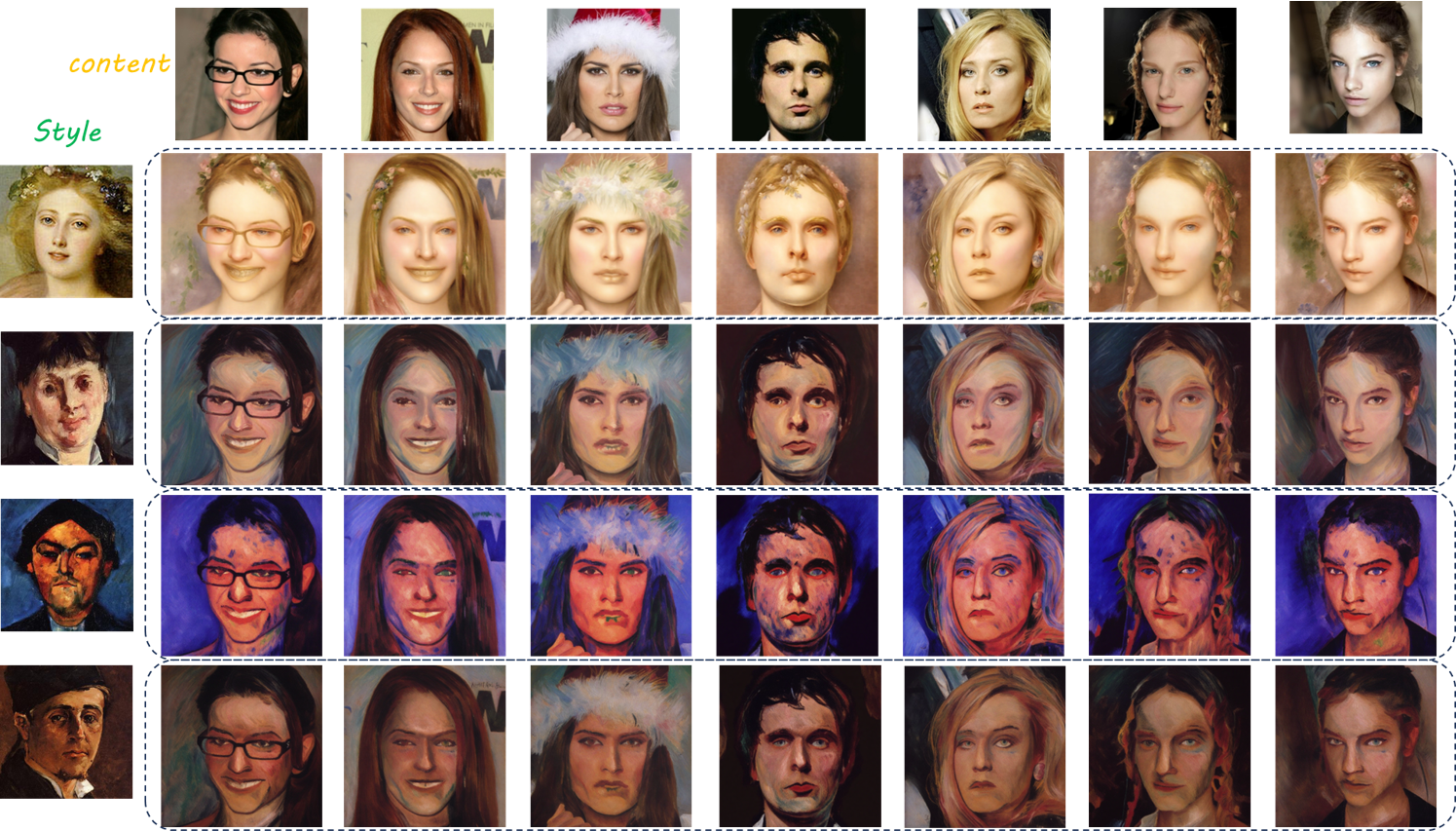}
     \caption{More Visualization Samples of the proposed \method$^*$.}
    \label{fig:add_vis_control_1}
\end{figure}
\section{Algorithm}
These algorithms detail our two-stage approach for geodesic-guided style transfer. Algorithm 1 presents the geodesic-guided style interpolation process (Section~\ref{sec:style_interp}). Starting from the linear interpolant, it applies a small number of rank-one refinement steps (Equation~\ref{eq:rank1_update}) that lower the discrete geodesic energy; it refines a single intermediate point and does not compute a full geodesic curve.

Algorithm 2 describes our style injection control mechanism, which operates in the attention space of the diffusion model. Starting from the linear query interpolant $(1-\beta)Q_c+\beta Q_s$, it applies the same truncated rank-one energy refinement to the query features (Section~\ref{sec:style_injection_control}).

Algorithm 3 integrates both components into a complete style transfer pipeline. The first stage refines the linearly interpolated content--style latent, and the second stage performs controlled style injection by refining the attention queries. Both stages use the same energy-minimization principle and the same stopping criterion. 
\begin{algorithm}[!h]
\caption{Geodesic-Guided Style Interpolation}
\begin{algorithmic}[1]
\REQUIRE Content image $x_c$, style image $x_s$, interpolation ratio $\delta$, learning rate $\lambda$, power-iteration steps $K_{\mathrm{PI}}$
\ENSURE Refined intermediate point $x_m$ (geodesic-guided point optimization)
\STATE Initialize $x_m = \delta x_c + (1-\delta)x_s$ 
\FOR{$i=1$ to $N_{iter}$}
    \STATE Initialize random unit vector $v$
    \FOR{$j=1$ to $K_{\mathrm{PI}}$}
        \STATE Compute $J_g v = \frac{\partial}{\partial a}g_d(x_m+av,t,c)|_{a=0}$
        \STATE Compute $w = \frac{\partial}{\partial x_m}\langle g_d(x_m,t,c), \mathrm{sg}(J_g v)\rangle$
        \STATE Normalize $v \leftarrow w/||w||$
    \ENDFOR
    \STATE Recompute $J_gv=\frac{\partial}{\partial a}g_d(x_m+av,t,c)|_{a=0}$ for the final $v$
    \STATE Compute $r_\delta=\delta g(x_c)+(1-\delta)g(x_s)-g(x_m)$ and $\eta_m=-2\alpha\,v(J_gv)^\top r_\delta$
    \STATE Update $x_m \leftarrow x_m - \lambda \eta_m$
    \STATE Stop early if the energy change is below $10^{-4}$
\ENDFOR
\STATE Return $x_m$
\end{algorithmic}
\end{algorithm}

\begin{algorithm}[!h]
\caption{Geodesic-based Style Injection Control}
\begin{algorithmic}[1]
\REQUIRE Content query $Q_c$, style query $Q_s$, interpolation parameter $\beta$, attention module $f$, learning rate $\lambda$, power-iteration steps $K_{\mathrm{PI}}$
\ENSURE Modified query features $Q_m$

\STATE Initialize mixed query: $Q_m \gets (1-\beta) Q_c + \beta Q_s$
\FOR{$i=1$ to $N1_{iter}$}
    \STATE $v \gets$ random unit vector
    \FOR{$j=1$ to $K_{\mathrm{PI}}$}
        \STATE $J_f v \gets \frac{\partial}{\partial a}f(Q_m + av)|_{a=0}$
        \STATE $w\gets\frac{\partial}{\partial Q_m}\langle f(Q_m),\mathrm{sg}(J_fv)\rangle$
        \STATE Normalize $v \gets w/||w||$
    \ENDFOR
    \STATE Recompute $J_fv=\frac{\partial}{\partial a}f(Q_m+av)|_{a=0}$ for the final $v$
    
    \STATE Compute rank-one gradient surrogate:
    \STATE $r_\beta\gets(1-\beta)f(Q_c)+\beta f(Q_s)-f(Q_m)$
    \STATE $\widehat{\nabla}_{Q_m} E_\beta \gets -2v(J_fv)^{\top}r_\beta$
    
    \STATE Update mixed query:
    \STATE $Q_m \gets Q_m - \lambda \widehat{\nabla}_{Q_m} E_\beta$
    \STATE Stop early if the energy change is below $10^{-4}$
\ENDFOR

\STATE Return $Q_m$
\end{algorithmic}
\end{algorithm}

\begin{algorithm}[!h]
\caption{Geodesic-Guided Style Transfer}
\begin{algorithmic}[1]
\REQUIRE Content image $x_c$, style image $x_s$, content-style ratio $\delta$, query interpolation ratio $\beta$, learning rate $\lambda$, power-iteration steps $K_{\mathrm{PI}}$
\ENSURE Stylized image
\STATE // Stage 1: Geodesic-guided style interpolation
\STATE Initialize $x_m = \delta x_c + (1-\delta)x_s$ 
\FOR{$i=1$ to $N_{iter}$}
    \STATE Initialize random unit vector $v$
    \FOR{$j=1$ to $K_{\mathrm{PI}}$}
        \STATE Compute $J_g v = \frac{\partial}{\partial a}g_d(x_m+av,t,c)|_{a=0}$
        \STATE Compute $w = \frac{\partial}{\partial x_m}\langle g_d(x_m,t,c), \mathrm{sg}(J_g v)\rangle$
        \STATE Normalize $v \leftarrow w/||w||$
    \ENDFOR
    \STATE Recompute $J_gv=\frac{\partial}{\partial a}g_d(x_m+av,t,c)|_{a=0}$ for the final $v$
    \STATE $r_\delta\leftarrow\delta g(x_c)+(1-\delta)g(x_s)-g(x_m)$
    \STATE $\eta_m\leftarrow-2\alpha\,v(J_gv)^\top r_\delta$
    \STATE Update $x_m \leftarrow x_m - \lambda \eta_m$
    \STATE Stop early if the energy change is below $10^{-4}$
\ENDFOR

\STATE // Stage 2: Style injection control (first SA layer of each transformer block, all 50 steps)
\STATE Extract queries $Q_c, Q_s$ from content and style images
\STATE Initialize $Q_m = (1-\beta) Q_c + \beta Q_s$
\FOR{$i=1$ to $N_{iter}$}
    \STATE Initialize random unit vector $v$
    \FOR{$j=1$ to $K_{\mathrm{PI}}$}
        \STATE Compute $J_f v = \frac{\partial}{\partial a}f(Q_m + av)|_{a=0}$
        \STATE Compute $w=\frac{\partial}{\partial Q_m}\langle f(Q_m),\mathrm{sg}(J_fv)\rangle$
        \STATE Normalize $v \leftarrow w/||w||$
    \ENDFOR
    \STATE Recompute $J_fv=\frac{\partial}{\partial a}f(Q_m+av)|_{a=0}$ for the final $v$
    \STATE $r_\beta\leftarrow(1-\beta)f(Q_c)+\beta f(Q_s)-f(Q_m)$
    \STATE $\widehat{\nabla}_{Q_m}E_\beta\leftarrow-2v(J_fv)^\top r_\beta$
    \STATE Update $Q_m \leftarrow Q_m - \lambda \widehat{\nabla}_{Q_m}E_\beta$
    \STATE Stop early if the energy change is below $10^{-4}$
\ENDFOR

\STATE // Final stylization
\STATE Replace query features with optimized $Q_m$ in attention layers
\STATE Stylized image using modified attention features
\end{algorithmic}
\end{algorithm}

\section{User Study} ~\label{app:user_study}

To complement the automatic metrics, we conducted a user study to assess the perceptual quality of the generated stylizations. We recruited 10 participants and presented each of them with 30 randomly selected content--style pairs. For each pair, the outputs of \method and three representative baselines (StyTR$^2$, StyleID, and StyleSSP) were shuffled and displayed side by side in an anonymous layout. Participants rated each result on a 1--5 Likert scale along four axes: \emph{style faithfulness} (how well the output captures the reference style), \emph{identity preservation} (how well the subject's identity is retained), \emph{visual quality} (overall aesthetic fidelity and absence of artifacts), and \emph{overall preference}. We report the mean score $\pm$ standard deviation for each axis in Table~\ref{tab:user_study}.

As shown in Table~\ref{tab:user_study}, \method obtains the highest mean scores among the four evaluated methods on all criteria, exceeding the strongest evaluated baseline (StyleID) by $0.42$, $0.45$, $0.34$, and $0.39$ on style faithfulness, identity preservation, visual quality, and overall preference, respectively. The standard deviations summarize the dispersion of the ratings; we do not interpret these descriptive results as a formal significance test.

\begin{table}[h]
    \centering
    \caption{User study results for the four evaluated methods (mean $\pm$ standard deviation, scale 1--5, higher is better).}
    \label{tab:user_study}
    \resizebox{\linewidth}{!}{
    \begin{tabular}{lcccc}
    \toprule
    Method & Style Faithfulness $\uparrow$ & Identity Preservation $\uparrow$ & Visual Quality $\uparrow$ & Overall Preference $\uparrow$ \\
    \midrule
    StyTR$^2$~\cite{deng2022stytr2}   & 3.62 $\pm$ 0.41 & 3.71 $\pm$ 0.38 & 3.68 $\pm$ 0.36 & 3.67 $\pm$ 0.39 \\
    StyleID~\cite{chung2024style}     & 3.89 $\pm$ 0.36 & 4.02 $\pm$ 0.31 & 3.94 $\pm$ 0.34 & 3.96 $\pm$ 0.33 \\
    StyleSSP~\cite{xu2025stylessp}    & 3.74 $\pm$ 0.38 & 3.85 $\pm$ 0.33 & 3.79 $\pm$ 0.35 & 3.80 $\pm$ 0.34 \\
    \midrule
    \method (Ours) & \textbf{4.31 $\pm$ 0.29} & \textbf{4.47 $\pm$ 0.24} & \textbf{4.28 $\pm$ 0.27} & \textbf{4.35 $\pm$ 0.26} \\
    \bottomrule
    \end{tabular}}
\end{table}

\section{Experiment Details}~\label{app:implement}

\subsection{Implementation Details:}
We use the official pretrained Stable Diffusion models, and the sample step is set to 50, the guidance scale is set to 3.0. The parameter $\delta$ is set to 0.9, $\alpha$ is set to 5, $\beta$ is set to 0.25 in our method.  The experiment is conducted on a single NVIDIA H800 GPU.  

\textbf{Map $g$ and Jacobian products.} $g$ is the complete frozen U-Net noise predictor at a fixed timestep and conditioning (Section~\ref{sec4.1}). Jacobian-vector and vector-Jacobian products are obtained by automatic differentiation through the composed computation graph $g_u\circ g_d$; $J_g$ is never materialized. The leading right singular vector is obtained by power iteration on $J_g^{\top}J_g$, and the update in Equation~\ref{eq:rank1_update} is applied for at most $K_{PI}$=10 iterations, with early stopping once the energy change drops below $10^{-4}$.

\textbf{Attention-level optimization.} The query optimization of Section~\ref{sec:style_injection_control} is applied to the first self-attention layer of each transformer block of the U-Net, at all spatial resolutions at which this layer occurs, and at all 50 DDIM sampling steps. Keys and values are taken from the style image and held fixed during the query optimization.

\subsection{Implementation of Baseline Methods}
For baseline methods, we use their official implementation and parameters.
For the two DiT-based models, Qwen-Image-Edit~\cite{wu2025qwen} and FLUX.2-Dev~\cite{labs2025flux}, we run the official inference code with the content and style images horizontally concatenated as a single two-panel input, paired with the following instruction prompt:
\begin{quote}
\small\itshape
``Two-panel image: left = style reference painting; right = a face photo to restyle. Edit only the right panel: repaint that portrait so it matches the left panel's medium, brushwork, color palette, and lighting, while preserving the same person, pose, framing, and facial identity. Leave the left panel unchanged as reference. Output the full wide image with the stylized result clearly visible on the right.''
\end{quote}
For the diffusion-based methods:
\begin{itemize}
    \item StyleSSP~\cite{xu2025stylessp}: A training-free diffusion-based style transfer method that enhances the sampling start point (the initial noise) to better preserve content. Its official implementation relies on ControlNet~\cite{zhang2023adding} and IP-Adapter~\cite{ye2023ip-adapter}, so in Table~\ref{tab:main_results} we compare it with \method$^*$, which uses the same additional controls.
    \item StyleID~\cite{chung2024style}: A training-free method that injects style by replacing the keys and values of the self-attention layers with those of the style image while preserving the content query (see the background above); \method builds on this injection scheme.
    \item DiffuseIT~\cite{kwon2022diffusion}: A diffusion-based image translation framework that disentangles style and content representations; we use it in its image-guided mode with the style image as reference, without additional training.
    \item InST~\cite{zhang2023inversion}: An inversion-based style transfer method that learns a textual embedding of the style reference and then uses it to guide the reverse diffusion process of the inverted content image.
    \item DiffStyle~\cite{jeong2024training}: A training-free method, originally named InjectFusion, that injects content features into the bottleneck ($h$-space) of the diffusion U-Net during the generation process of the style image.
\end{itemize}
and four conventional style transfer methods:
\begin{itemize}
    \item AesPA-Net~\cite{hong2023aespa}: A feed-forward arbitrary style transfer network that uses the pattern repeatability of the style image to adapt its attention-based feature transfer, aiming at aesthetically pleasing results.
    \item CAST~\cite{zhang2022domain}: An arbitrary style transfer method that learns domain-enhanced style representations with contrastive learning.
    \item StyTR$^2$~\cite{deng2022stytr2}: An improved transformer-based framework that treats style transfer as a sequence-to-sequence task, using an encoder-decoder architecture to capture both global and local style patterns.
    \item ArtFlow~\cite{an2021artflow}: A style transfer method based on reversible neural flows that provides a bias-free transformation, ensuring precise content preservation by learning invertible mappings between content and style feature distributions.
\end{itemize}
\subsection{Computation Cost} \label{app:compute_cost}
We provide detailed analysis of computational requirements and compare our method with state-of-the-art baselines. The analysis demonstrates that our approach achieves superior quality-efficiency trade-off among training-free methods.
\rev{
\textbf{Timing Comparison.} Table~\ref{tab:timing} presents per-image inference time measurements for 512$\times$512 images on a single NVIDIA H800 GPU. Our method processes one image in 9.71 seconds, which is dramatically faster than training-based diffusion methods: 23$\times$ faster than DiffuseIT (226.82s) and 7$\times$ faster than InjectFusion (68.24s). Among training-free methods, StyleID achieves faster inference (3.97s, 2.4$\times$ faster than ours), but delivers substantially lower quality across all metrics as discussed below.
}
\begin{table}[h]
    \centering
    \caption{Computational comparison on 512$\times$512 images (NVIDIA H800 GPU).}
    \resizebox{0.8\linewidth}{!}{
    \begin{tabular}{lccc}
    \toprule
    Method & Inference Time (s) & Training Required & Speedup vs. Ours \\
    \midrule
        DiffuseIT & 226.82 & No & 0.04$\times$ (23$\times$ slower) \\
    DiffStyle & 68.24 & No & 0.14$\times$ (7$\times$ slower) \\
    StyleID & 3.97 & No & 2.4$\times$ (faster) \\
     StyleSSP & 6.56 & No & 1.5× (faster) \\
    \midrule
    Ours & 9.71 & No & 1.0$\times$ \\
    \bottomrule
    \end{tabular}
    }
    \label{tab:timing}
\end{table}

\textbf{Training Requirements.} A critical advantage of our approach is that it requires no training or fine-tuning. We use pretrained Stable Diffusion models directly without modification. In contrast, DiffuseIT and InjectFusion require training on style-specific datasets, which takes several hours on multiple GPUs and must be repeated for each new style domain. This training overhead severely limits flexibility for diverse style transfer applications. Our training-free nature enables immediate application to arbitrary new styles without any preparation beyond having the style reference image.

\rev{\textbf{Computational Efficiency.} Our geodesic optimization achieves efficiency through three key designs: (1) automatic differentiation computes Jacobian--vector and vector--Jacobian products without materializing the full Jacobian matrix, (2) power iteration provides the rank-one Jacobian approximation, with early stopping when the energy change drops below $10^{-4}$, and (3) activation checkpointing recomputes forward passes during backpropagation rather than storing all intermediate activations. These optimizations make the geodesic computation practical while maintaining superior quality.}

\textbf{Quality-Efficiency Trade-off.} The comparison with StyleID is particularly revealing. While StyleID is 2.4$\times$ faster, our method achieves substantially better quality (see Table~\ref{tab:main_results}): identity preservation (ID: 0.1237 vs. 0.1618 on WikiArt), perceptual distance (LPIPS: 0.2616 vs. 0.3785), and content similarity (CS: 0.5531 vs. 0.4493). The modest computational overhead of geodesic navigation is justified by significant quality improvements across all evaluation metrics. This represents the fundamental advantage of respecting manifold geometry: geodesic paths produce better results than simple interpolation schemes.

In summary, our method achieves the optimal balance for training-free style transfer: no domain-specific training required, practical inference times suitable for content creation workflows (approximately 370 images per hour), and superior quality compared to alternative training-free approaches. The 9.71-second processing time is entirely practical for professional applications while delivering state-of-the-art quality.

\subsection{Performance on Different Base Model}
To evaluate the robustness and generalization capability of our \method, we conduct experiments using different versions of Stable Diffusion (SD) as the base model. We compare our approach with StyleID across SD v1.5, v2.0, and v2.1. Table~\ref{tab:sd_results} presents a comprehensive comparison of the performance metrics across these different base models.

\begin{table*}[h]
    \centering
    \small
    \caption{Style transfer results of StyleID and \method with different Stable Diffusion base models, using WikiArt and MetFace as style references.}
    \resizebox{0.9\textwidth}{!}{
    \begin{tabular}{cc|cccc|cccc}
      % Method  & \multicolumn{8}{c|}{WikiArt} & \multicolumn{8}{c}{MetFace} \\
      \toprule
      \multirow{2}{*}{Base Model} &\multirow{2}{*}{Method}& \multicolumn{4}{c|}{\textbf{WikiArt}} & \multicolumn{4}{c}{\textbf{MetFace}} \\
       ~ & ~ & ArtFID ($\downarrow$) & LPIPS ($\downarrow$) & ID ($\downarrow$) & CS ($\uparrow$) &  ArtFID ($\downarrow$) & LPIPS ($\downarrow$) & ID ($\downarrow$) & CS ($\uparrow$)\\
       \midrule
        \multirow{2}{*}{\textbf{SD v1.5}} & StyleID & 20.6744& 0.3878 & 0.1273& 0.4679&\textbf{25.3002 }& 0.4044 &  0.0952 & 0.4742\\
        ~  & \method & \textbf{20.2082} &\textbf{ 0.2737} & \textbf{0.1242 }& \textbf{0.5502} & 26.5815 & \textbf{0.2436} & \textbf{0.0626 }& \textbf{0.5941} \\
        \midrule
        \multirow{2}{*}{\textbf{SD v2.0}} & StyleID & 20.8419 & 0.3850 & 0.1271 &  0.4715  & 25.3235 & 0.3992 &  0.0915 &   0.4817 \\
        ~  & \method & \textbf{20.0907} & \textbf{0.2508} & \textbf{0.1177} & \textbf{0.5758} & \textbf{24.5590} & \textbf{0.2580} & \textbf{0.0649 }& \textbf{0.6033} \\
        \midrule
        \multirow{2}{*}{\textbf{SD v2.1}} & StyleID & 25.3337 & 0.4200 & 0.1351 & 0.4244 & 27.9503 & 0.5190 &  0.1234 &  0.4077 \\
        ~  & \method & \textbf{20.8076} & \textbf{0.2527 }&  \textbf{0.1175} &\textbf{ 0.5783} & \textbf{26.7607} & \textbf{0.2601} & \textbf{0.0659 }&   \textbf{0.6016}\\
        \bottomrule
    \end{tabular}
    }
    \label{tab:sd_results}
\end{table*}
\vspace{-8pt}

The results in Table~\ref{tab:sd_results} show that the behavior of \method is consistent across SD versions. For the WikiArt dataset, our approach achieves better performance across all metrics, with notable improvements in LPIPS scores (reducing from 0.3878 to 0.2737 in SD v1.5) and content similarity (CS $>$ 0.55). On the MetFace dataset, \method achieves better LPIPS, ID, and CS for all versions and better ArtFID for SD v2.0 and v2.1, while StyleID attains a better ArtFID with SD v1.5. Notably, our \method maintains consistent performance across all SD versions, demonstrating its robustness and generalizability regardless of the underlying diffusion model architecture.

\subsection{General Style Transfer} \label{app:general}

\rev{While our main paper focuses on facial style transfer, our method is fundamentally domain-agnostic. The geodesic framework relies purely on the learned manifold geometry of diffusion models without any face-specific priors or constraints. To validate this generality, we conduct comprehensive experiments across three diverse non-face domains: art-to-art transfer, object stylization, and scene stylization. To validate the domain-agnostic nature of our method, we conduct experiments on diverse non-facial content. We select content images from MS-COCO, which provides natural images across various categories including objects, animals, and complex scenes. Style reference images are randomly sampled from WikiArt~\cite{tan2018improved}, spanning diverse artistic movements. This setup tests whether our geodesic framework generalizes beyond facial images to arbitrary content-style pairs.}

\begin{figure*}[h]
    \centering
    \begin{subfigure}{\linewidth}
        \includegraphics[width=\linewidth]{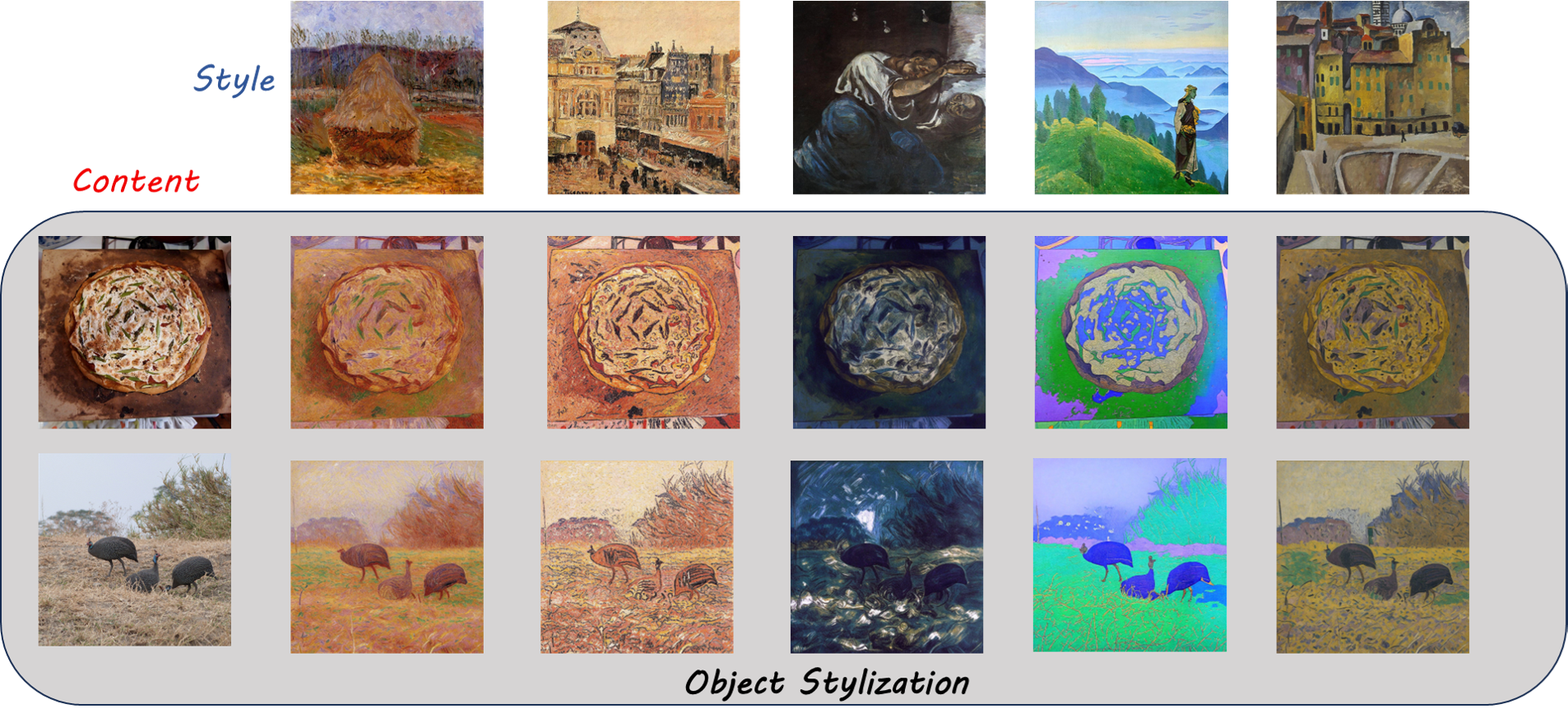}
        \caption{Object Stylization.}
        \label{fig:object}
    \end{subfigure}
    \begin{subfigure}{\linewidth}
        \includegraphics[width=\linewidth]{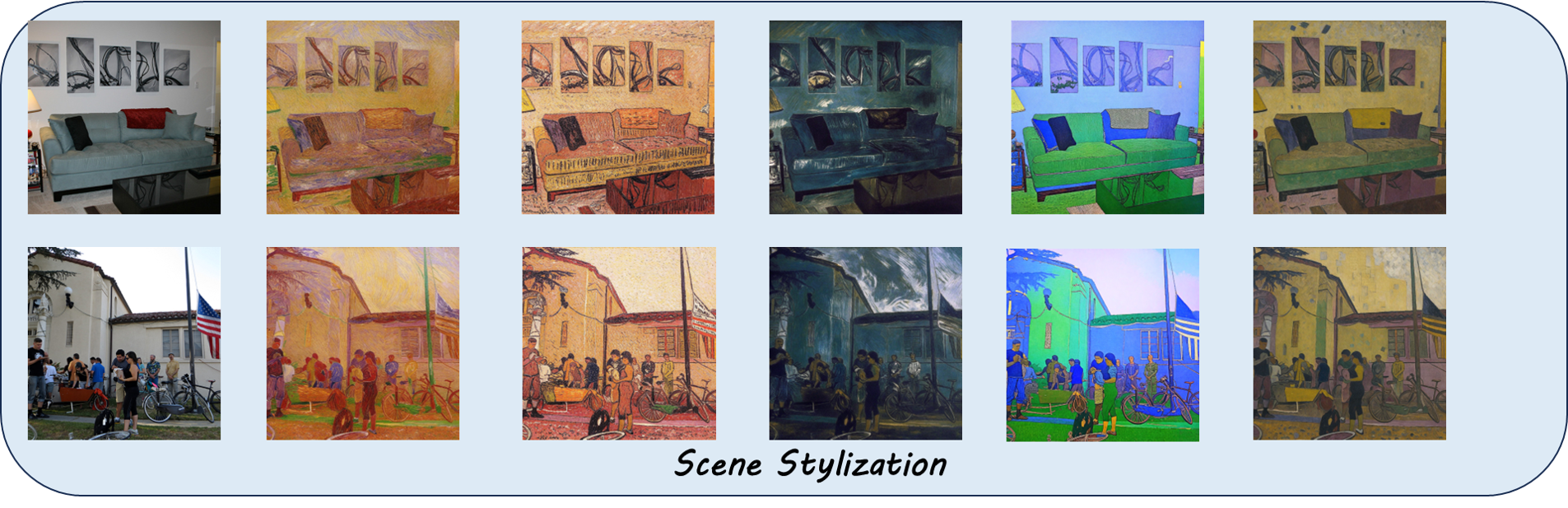}
        \caption{Scene Stylization.}
        \label{fig:scene}
    \end{subfigure}
    \begin{subfigure}{\linewidth}
        \includegraphics[width=\linewidth]{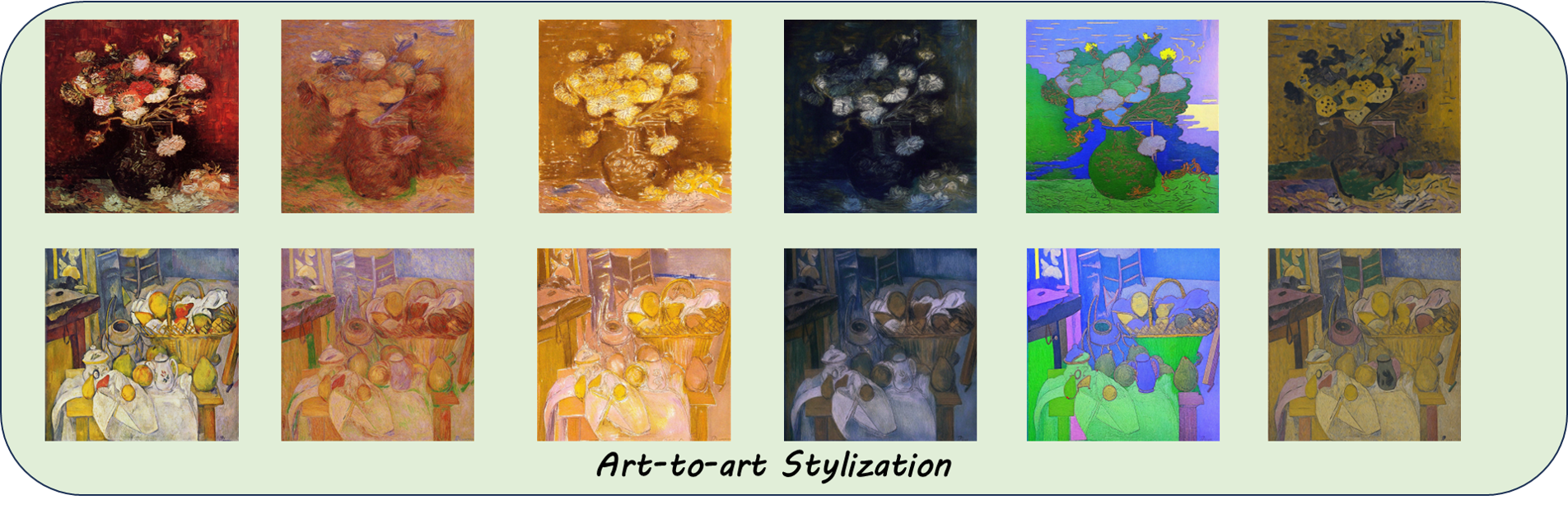}
        \caption{Art-to-art Stylization.}
        \label{fig:art2art}
    \end{subfigure}
    \caption{The style transfer results of object stylization, scene stylization and art-to-art transfer.}
    \label{fig:general_trans}
\end{figure*}

\textbf{Object Stylization.} We evaluate style transfer on objects with clear semantic structure, including animals (birds), and everyday objects (pizza). As shown in Figure~\ref{fig:general_trans} (top row), the geodesic paths preserve object structure and identity while applying artistic textures and color transformations. Object shapes remain intact while adopting the style's visual characteristics, validating that our manifold-based approach respects semantic boundaries beyond facial features.

\textbf{Scene Stylization.} We test on complex scenes from MS-COCO. Scene stylization is particularly challenging due to multiple objects, varying scales, and complex spatial relationships. Figure~\ref{fig:general_trans} (middle row) demonstrates that our method maintains spatial coherence and layout consistency while transferring style. The geodesic paths naturally preserve the scene's compositional structure--horizon lines, building arrangements, foreground-background relationships--while applying artistic transformations.

\textbf{Art-to-Art Transfer.} We test style transfer between artistic paintings without facial content, sampling painting pairs from WikiArt spanning diverse artistic movements. As shown in Figure~\ref{fig:general_trans} (bottom row), our method successfully captures and transfers artistic characteristics-brush stroke patterns, color palettes, and texture styles--while maintaining the content painting's compositional structure and spatial layout.

\subsection{Case Study} \label{app:case_study}
\begin{figure*}[!h]
    \centering
    \begin{minipage}[b]{0.48\linewidth}
    \includegraphics[width=\linewidth]{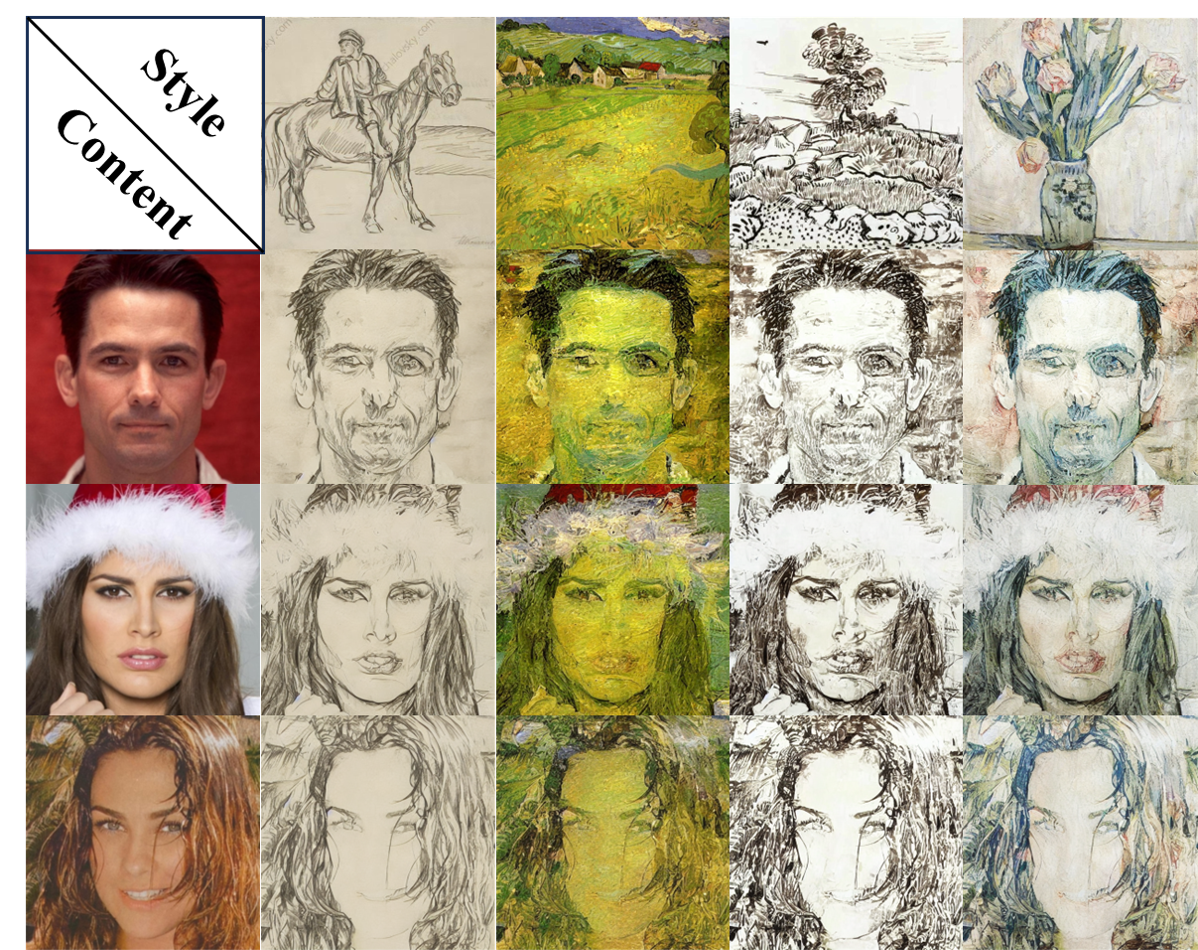}
    \caption{Style transfer results using non-facial artistic references as style images.}
    \label{fig:case_2}
    \end{minipage}
    \hspace{4pt}
    \begin{minipage}[b]{0.48\linewidth}
            \centering
    \includegraphics[width=\linewidth]{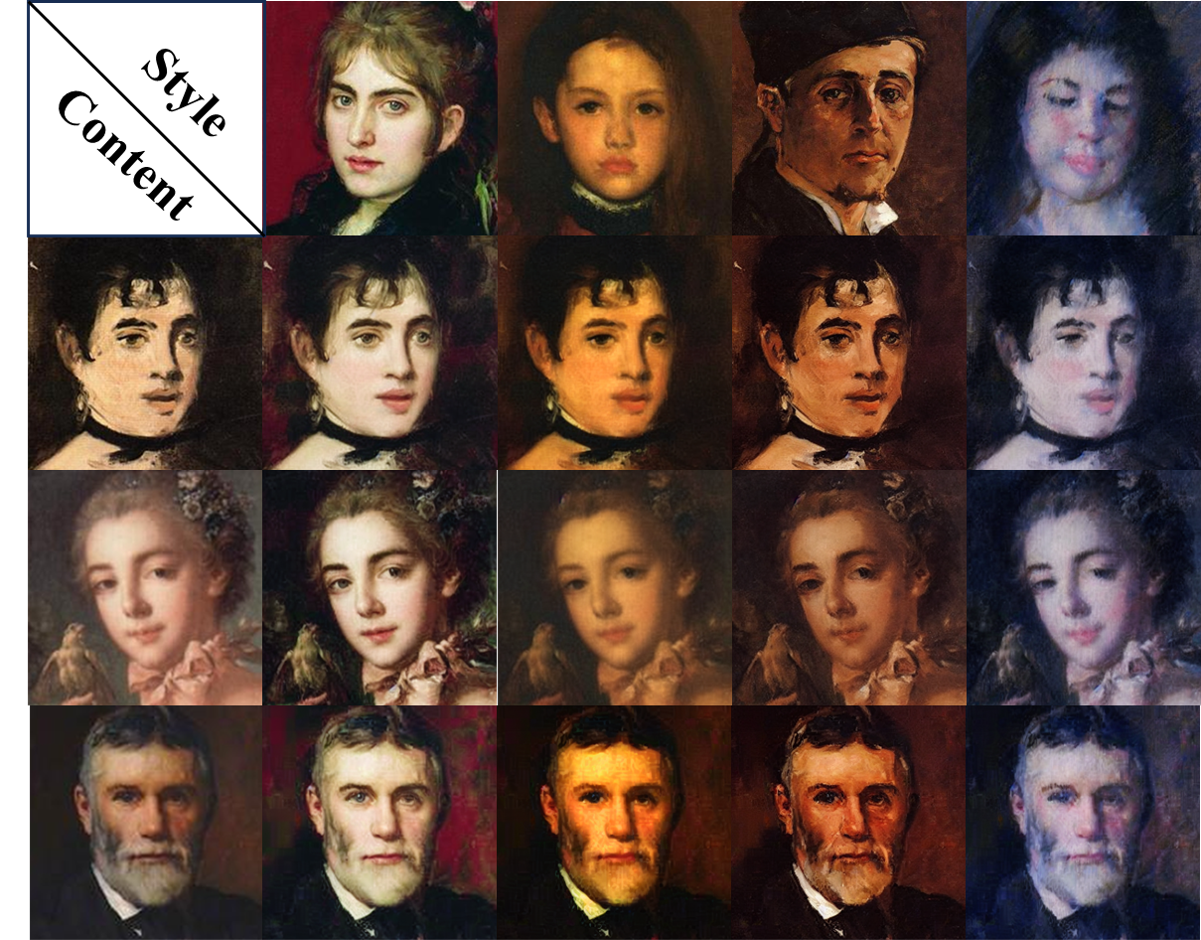}
    \caption{Style transfer results between different artistic portrait images.}
    \label{fig:case_1}
    \end{minipage}
\end{figure*}
\textbf{Using Non-Facial Art References}
We evaluate our \method's capability to transfer styles from non-facial artistic references. As shown in Figure~\ref{fig:case_2}, our approach successfully extracts and applies artistic elements (color palettes, brush strokes, textures) from non-portrait artworks while preserving facial identity and human features. The results demonstrate that our \method can effectively handle diverse artistic references beyond portraits, from abstract paintings to landscapes, while maintaining content fidelity.
%While our method primarily focuses on artistic portrait style transfer, we also explore its capability to handle non-facial artistic references. This case study demonstrates how our approach can effectively extract and apply artistic styles from various non-portrait artworks while maintaining facial identity and realistic human features. We test our method using reference images ranging from abstract paintings to landscapes, showing its versatility in style extraction and application.

%Figure~\ref{fig:case_2} presents examples where we use non-facial artworks as style references. The results show that our method successfully captures the color palettes, brush stroke patterns, and artistic elements from these diverse references while ensuring the output maintains recognizable human features. This ability to work with non-facial art references significantly expands the creative possibilities of our method, allowing users to draw inspiration from a broader range of artistic sources.

\textbf{Artistic style to artistic style}
Our \method can effectively transfer artistic styles between different artistic images while preserving the content and identity of the source image. Figure~\ref{fig:case_1} shows an example where we transfer the style from one artistic portrait to another. The results demonstrate that our \method successfully captures and transfers the distinctive artistic characteristics while maintaining the facial features and overall composition of the content image. This capability is particularly valuable for artists and content creators who wish to experiment with different artistic styles while ensuring the subject remains recognizable.

\subsection{Effect of Inference Step} \label{app:inference_step}
\begin{figure*}[h]
    \centering
    \includegraphics[width=\textwidth]{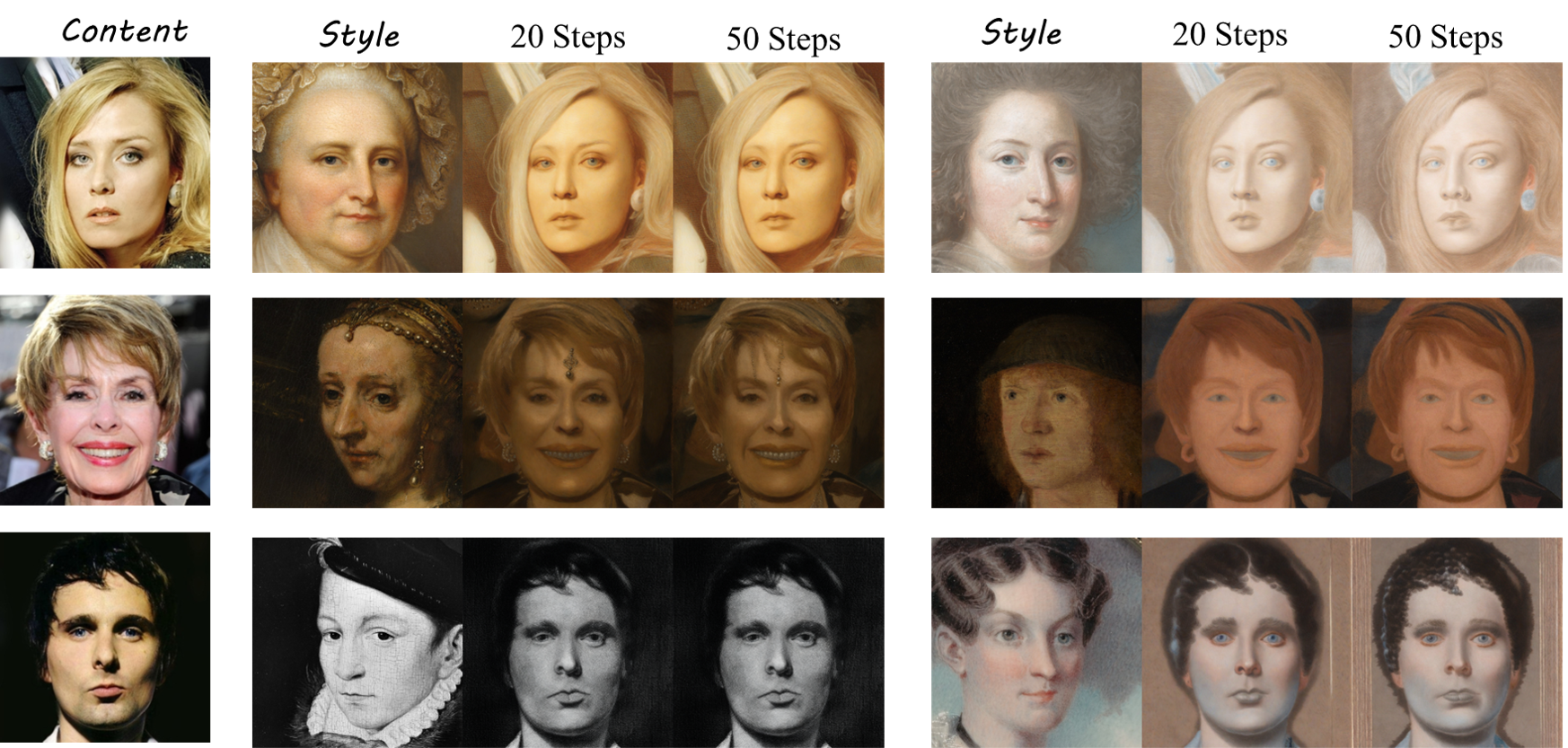}
    \caption{Effect of the number of sampling steps (20 vs.\ 50) on style transfer performance, using MetFace as style references.}
    \label{fig:inference_step}
\end{figure*}

\rev{We evaluate the effect of inference steps on the performance of our \method. Figure~\ref{fig:inference_step} shows a comparison between using 20 steps versus our standard 50 steps. The 20-step version shows only a modest decrease in style fidelity and identity preservation, while content similarity remains largely stable. This demonstrates that our method offers flexible quality-speed trade-offs suitable for different application scenarios - users can choose 20 steps for near-real-time applications or 50 steps for highest quality results.}

\begin{figure*}[!h]
    \centering
    \begin{subfigure}{0.11\linewidth}
        \includegraphics[width=\linewidth]{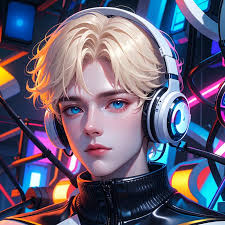}
        \caption{Style}
    \end{subfigure}
    \hfill
    \begin{subfigure}{0.11\linewidth}
        \includegraphics[width=\linewidth]{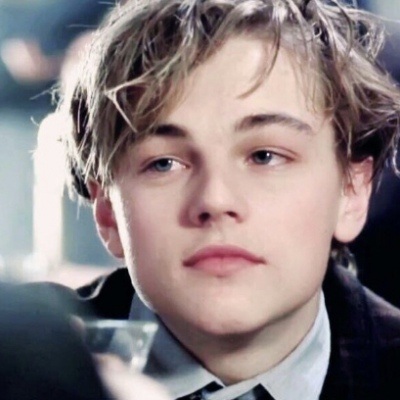}
        \caption{Content}
    \end{subfigure}
    \hfill
\begin{subfigure}{0.11\linewidth}
        \includegraphics[width=\linewidth]{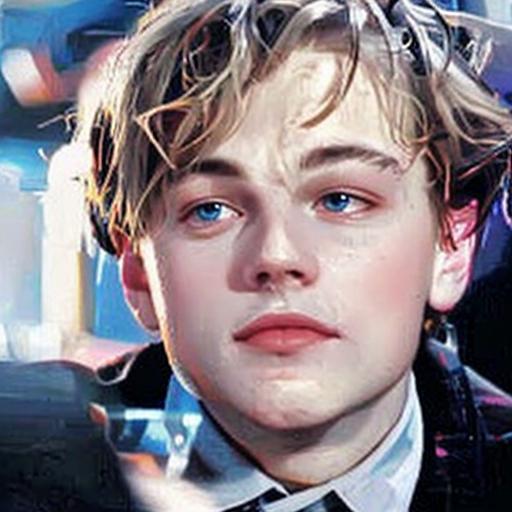}
        \caption{$\delta=0.$}
    \end{subfigure}
    \hfill
    \begin{subfigure}{0.11\linewidth}
        \includegraphics[width=\linewidth]{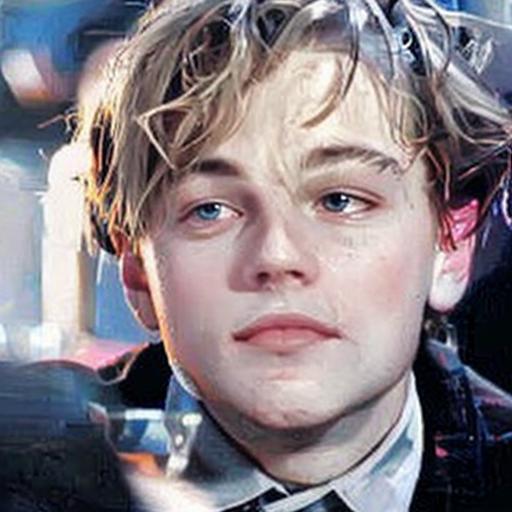}
        \caption{$\delta=0.2.$}
    \end{subfigure}
    \hfill
    \begin{subfigure}{0.11\linewidth}
        \includegraphics[width=\linewidth]{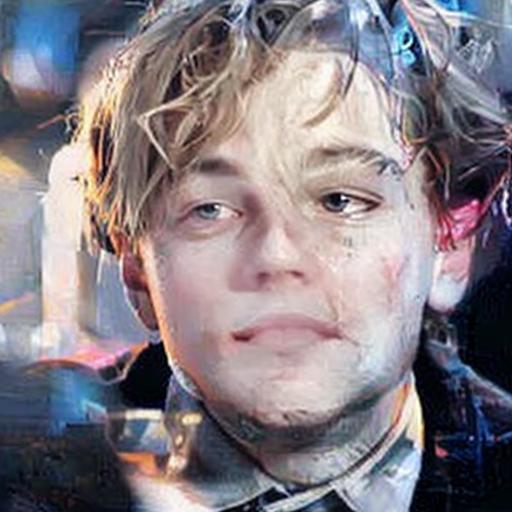}
        \caption{$\delta=0.4.$}
    \end{subfigure}
    \hfill
    \begin{subfigure}{0.11\linewidth}
        \includegraphics[width=\linewidth]{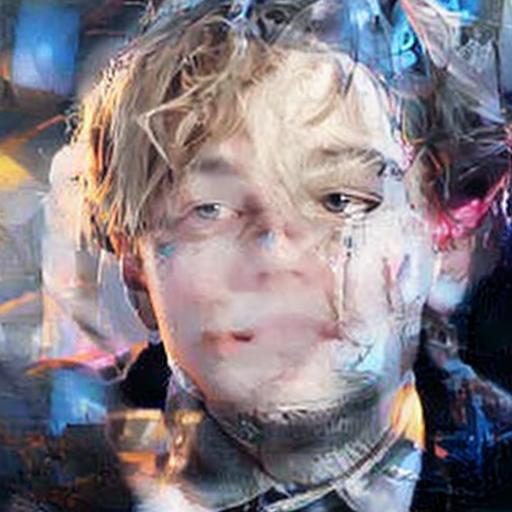}
        \caption{$\delta=0.6.$}
    \end{subfigure}
    \hfill
    \begin{subfigure}{0.11\linewidth}
        \includegraphics[width=\linewidth]{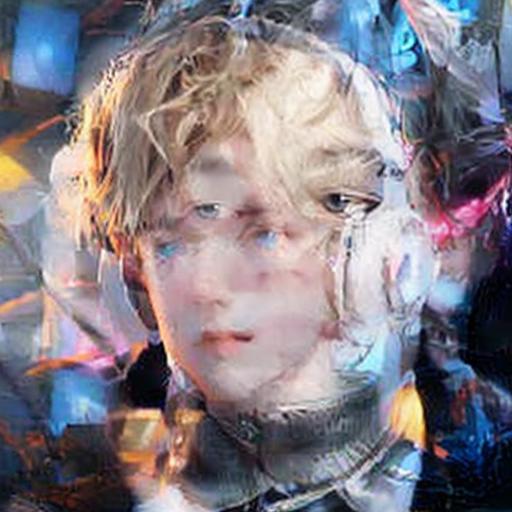}
        \caption{$\delta=0.8.$}
    \end{subfigure}
    \hfill
    \begin{subfigure}{0.11\linewidth}
        \includegraphics[width=\linewidth]{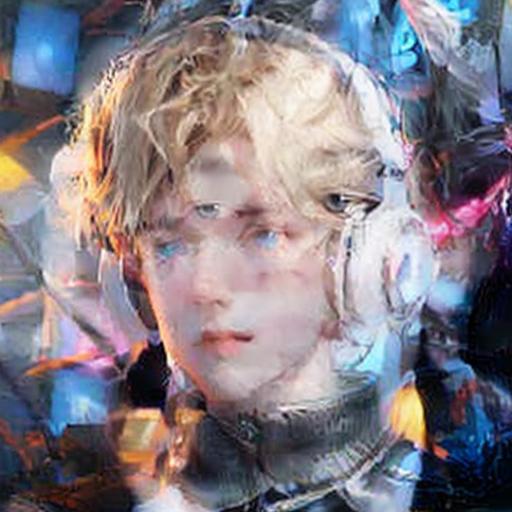}
        \caption{$\delta=1.0.$}
    \end{subfigure}
    \caption{Linear baseline interpolation.}
    \label{fig:linear_vis}
\end{figure*}

\subsection{Linear Interpolation Visualization}
Figure~\ref{fig:linear_vis} shows the results of purely linear interpolation between content and style latents, i.e., the initialization $x_m=\delta x_c+(1-\delta)x_s$ without the energy-based refinement. The transitions are not smooth, the style is transferred poorly, identity is not well preserved, and blur and sharpening artifacts are visible.
\iffalse
\begin{figure*}[!h]
    \centering
    \iffalse
    \begin{subfigure}{0.12\linewidth}
        \includegraphics[width=\linewidth]{sec/figures/commic_img2.jpg}
        \caption{Style}
    \end{subfigure}
    \begin{subfigure}{0.12\linewidth}
        \includegraphics[width=\linewidth]{sec/figures/leangnaduo.jpeg}
        \caption{Content}
    \end{subfigure}
    \begin{subfigure}{0.12\linewidth}
        \includegraphics[width=\linewidth]{sec/figures/lambda=0.2.png}
        \caption{$\lambda=0.2.$}
    \end{subfigure}
    \begin{subfigure}{0.12\linewidth}
        \includegraphics[width=\linewidth]{sec/figures/lambda=0.4.png}
        \caption{$\lambda=0.4.$}
    \end{subfigure}
    \begin{subfigure}{0.12\linewidth}
        \includegraphics[width=\linewidth]{sec/figures/lambda=0.6.png}
        \caption{$\lambda=0.6.$}
    \end{subfigure}
    \begin{subfigure}{0.12\linewidth}
        \includegraphics[width=\linewidth]{sec/figures/lambda=0.8.png}
        \caption{$\lambda=0.8.$}
    \end{subfigure}
    \begin{subfigure}{0.12\linewidth}
        \includegraphics[width=\linewidth]{sec/figures/lambda=1.0.png}
        \caption{$\lambda=1.0.$}
    \end{subfigure}
    \fi 
    \includegraphics[width=\linewidth]{sec/figures/style_interp.png}
    \caption{Linear baseline interpolation.}
    \label{fig:linear_vis}
\end{figure*}
\fi

\section{Visualization} ~\label{app:visualization}

\subsection{More Visualization Results}
\begin{figure*}[!h]
    \centering
    \includegraphics[width=\linewidth]{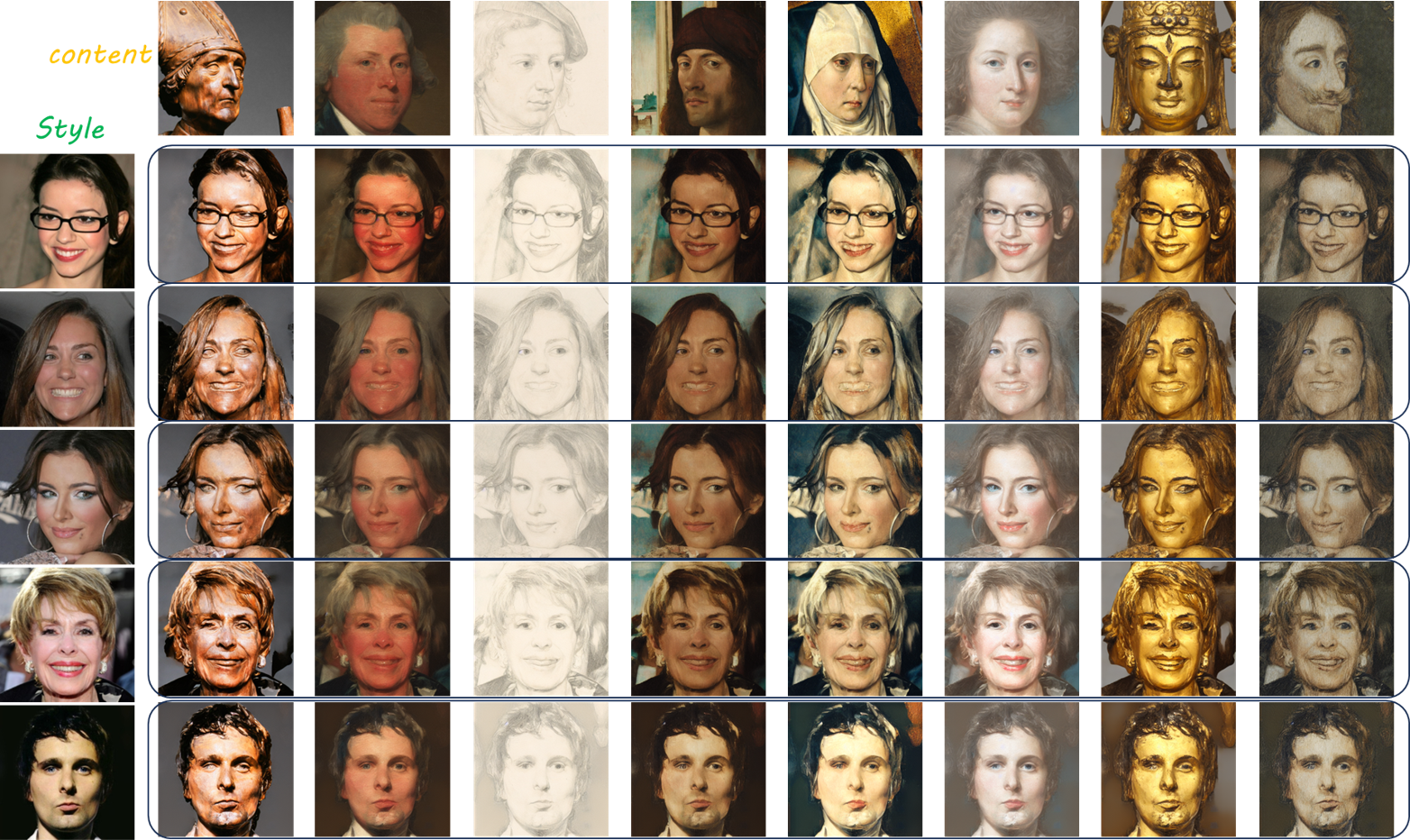}
    \caption{More visualization results showing style transfer examples with different content and style image pairs.}
    \label{fig:supp_visual_main}
\end{figure*}

In this section, we present additional visualization results to demonstrate the effectiveness and versatility of our method. Figure~\ref{fig:supp_visual_main} shows additional style transfer examples with \method, Figure~\ref{fig:add_vis_control_1} and Figure~\ref{fig:add_vis_control} shows additional style transfer examples with \method$^*$,  illustrating how our approach successfully transfers artistic characteristics while preserving facial identity and semantic structure. The results showcase smooth transitions between content and style domains, with consistent quality across diverse artistic styles and facial features. These visualizations complement the main experimental results by providing a more comprehensive view of our method's capabilities in handling various artistic references and content images.

\subsection{More Comparison Results}
\begin{figure*}[!h]
    \centering
    \includegraphics[width=\linewidth]{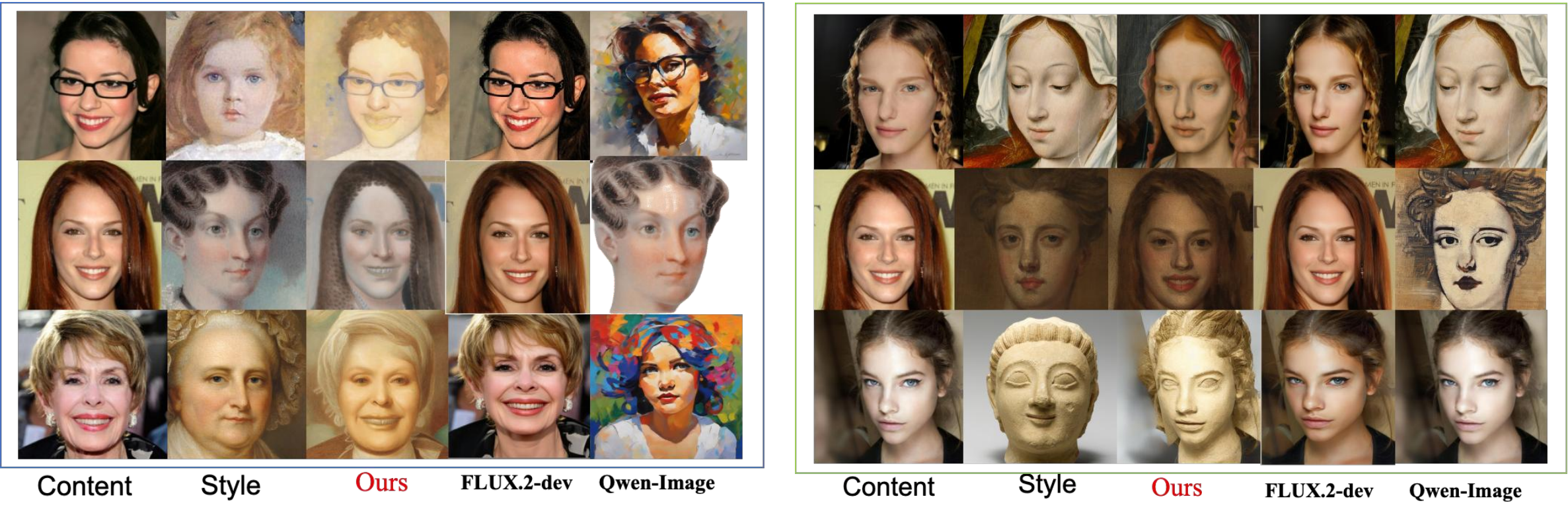}
    \caption{Qualitative comparison of style transfer results against DiT-based editing models.}
    \label{fig:add_comp_dit}
\end{figure*}

\begin{figure*}[!h]
    \centering
    \includegraphics[width=\linewidth]{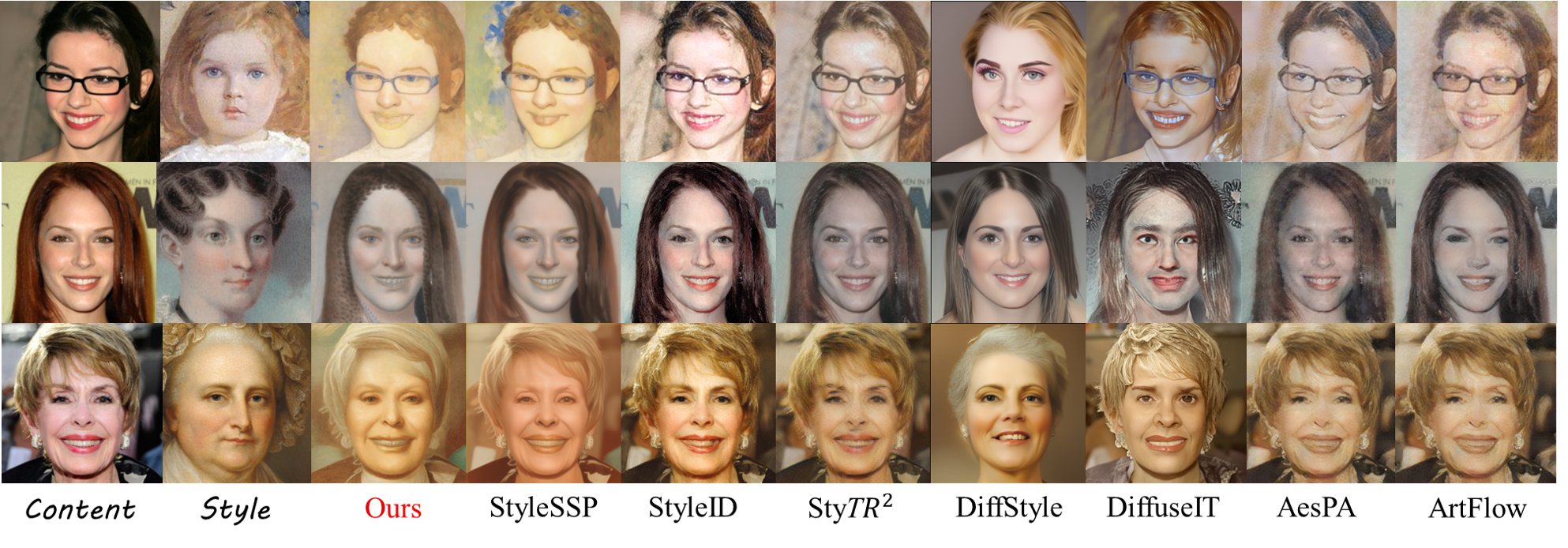}
    \caption{Qualitative comparison of style transfer results against diffusion-based and conventional baselines.}
    \label{fig:add_comparison}
\end{figure*}

Owing to space constraints in the main paper, we split the qualitative comparisons across two figures. Figure~\ref{fig:add_comp_dit} compares our method against recent DiT-based editing models, while Figure~\ref{fig:add_comparison} reports comparisons against diffusion-based and conventional style transfer baselines. Together, these results further demonstrate the superiority of our approach in balancing faithful style transfer with identity preservation.

\subsection{Zoom-in Visualization}
\begin{figure*}[!h]
    \centering
    \begin{subfigure}[b]{\linewidth}
        \includegraphics[width=\linewidth]{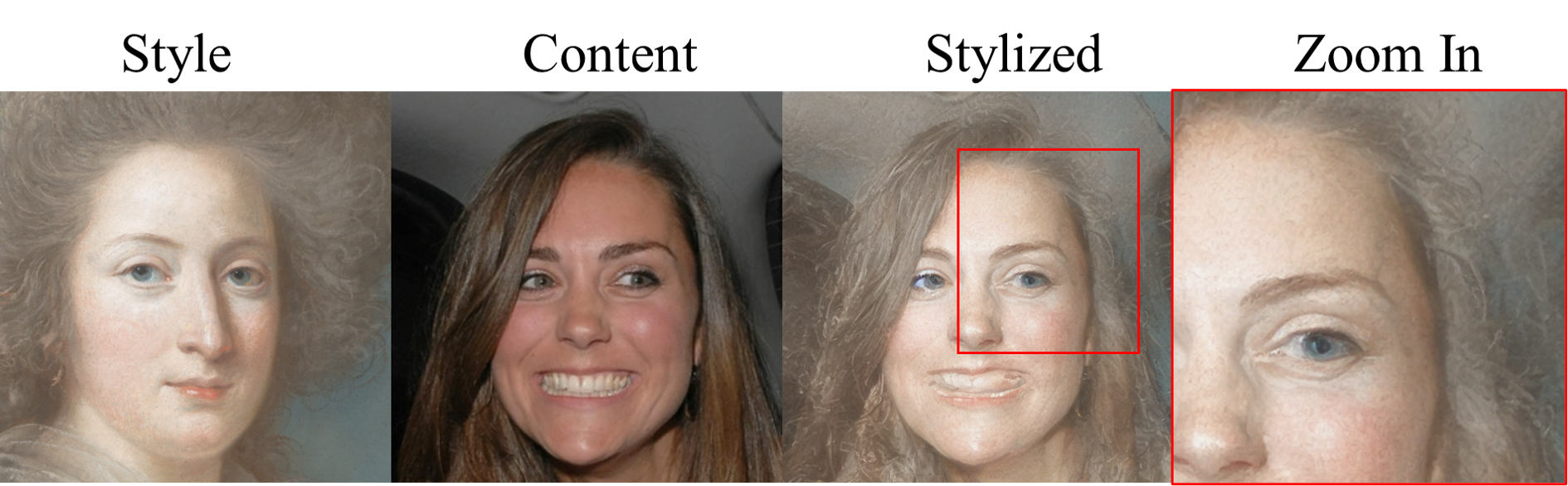}
    \end{subfigure}
    
    \begin{subfigure}[b]{\linewidth}
        \includegraphics[width=\linewidth]{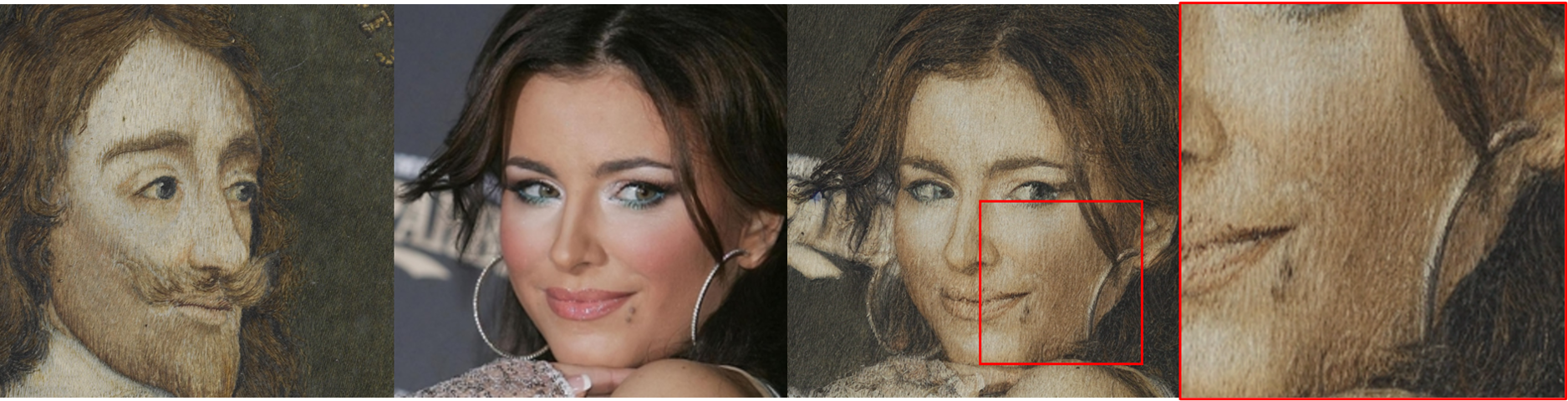}
    \end{subfigure}
    
    \begin{subfigure}[b]{\linewidth}
        \includegraphics[width=\linewidth]{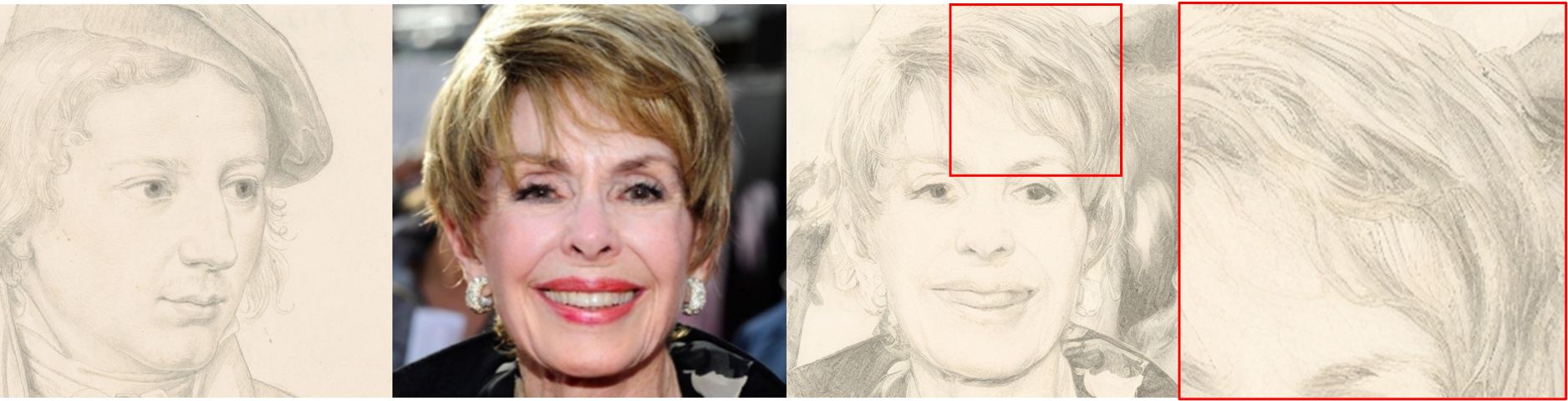}
    \end{subfigure}
    
     \begin{subfigure}[b]{\linewidth}
        \includegraphics[width=\linewidth]{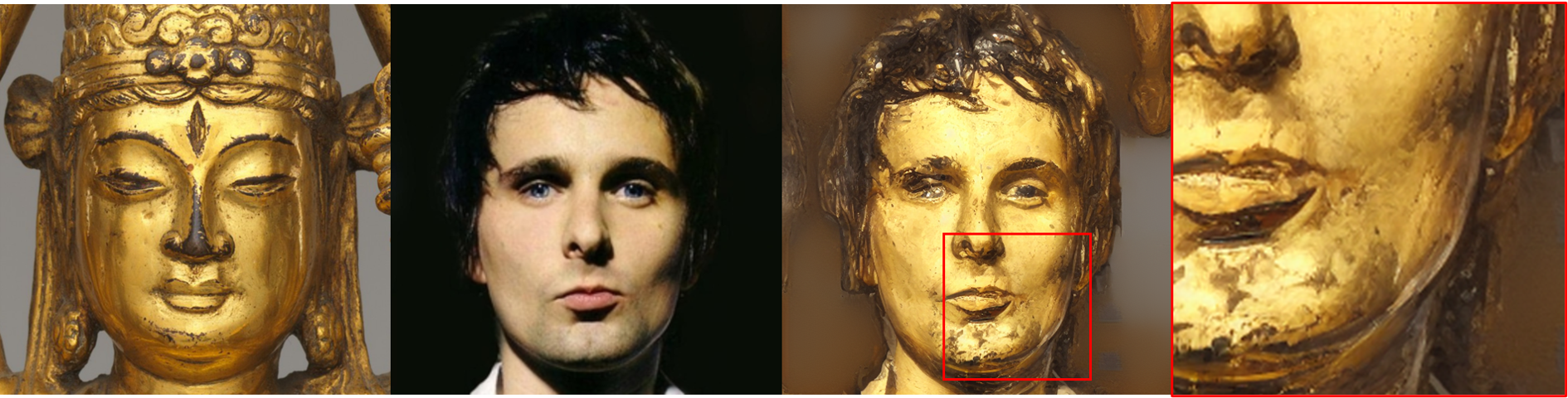}
    \end{subfigure}
    \caption{Zoom-in visualization comparing original and stylized image regions. The detailed views highlight our method's ability to preserve facial features while incorporating artistic elements.}
    \label{fig:zoom}
\end{figure*}

To provide a detailed view of how our method handles fine-grained style transfer, we present zoom-in visualizations that highlight specific regions of interest. Figure~\ref{fig:zoom} shows magnified portions of our stylization results, focusing on facial features, textures, and artistic elements. These detailed views demonstrate how our method successfully preserves important facial characteristics while incorporating artistic style elements at different scales.

\subsection{More Visualization of artistic to artistic style transfer}
\begin{figure*}[!h]
    \centering
    \includegraphics[width=\linewidth]{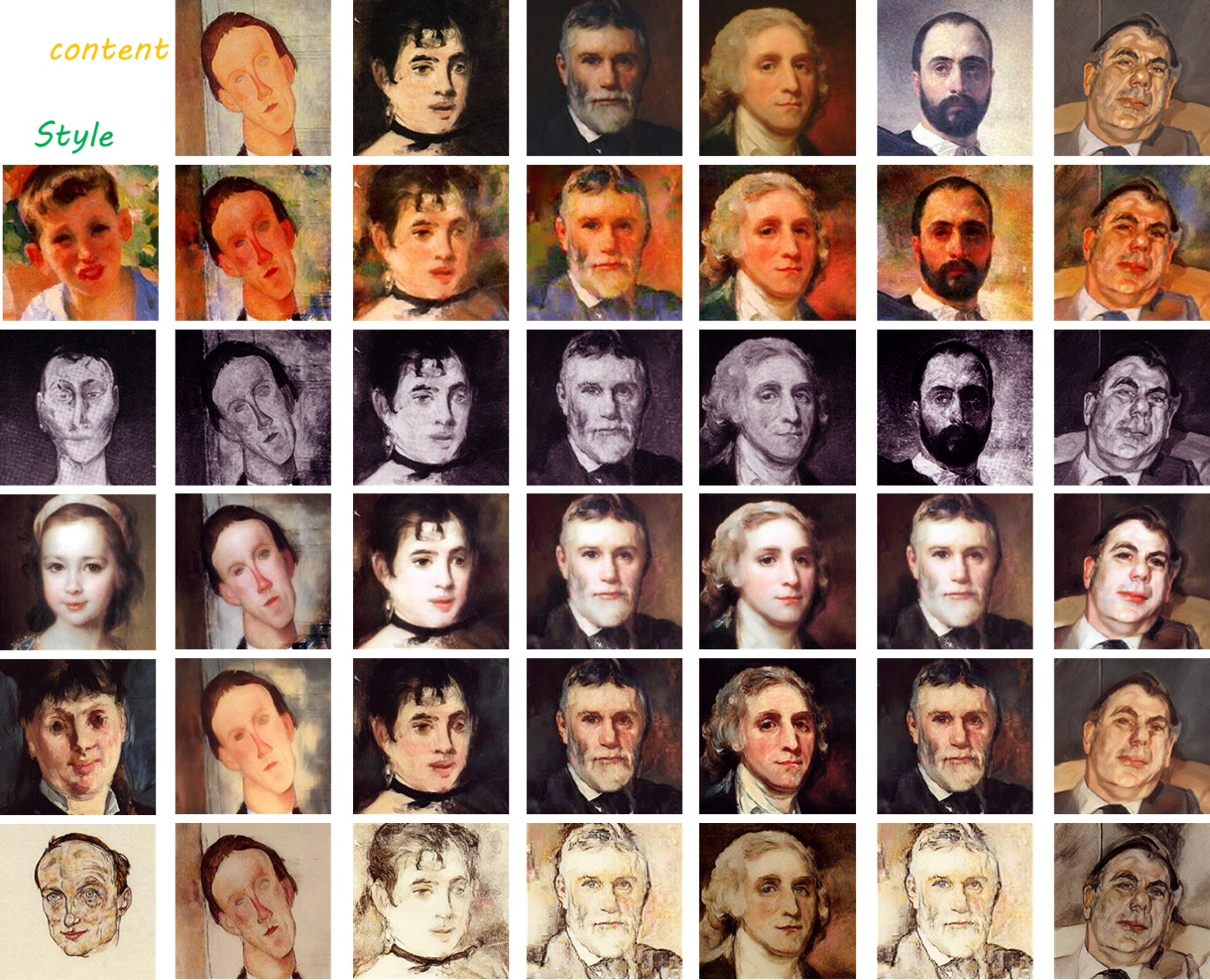}
    \caption{Additional artistic-to-artistic style transfer results. Each row shows a different example with content image, style reference, and our stylization result.}
    \label{fig:app_art2art}
\end{figure*}

We provide additional visualization results of artistic-to-artistic style transfer to demonstrate our method's capability in handling diverse artistic styles. As shown in Figure~\ref{fig:app_art2art}, our method successfully transfers styles between different artistic portraits while preserving the key characteristics of both content and style images.

\subsection{More Visualization of non-face artistic to realistic style transfer}
\begin{figure*}[!h]
    \centering
    \includegraphics[width=\linewidth]{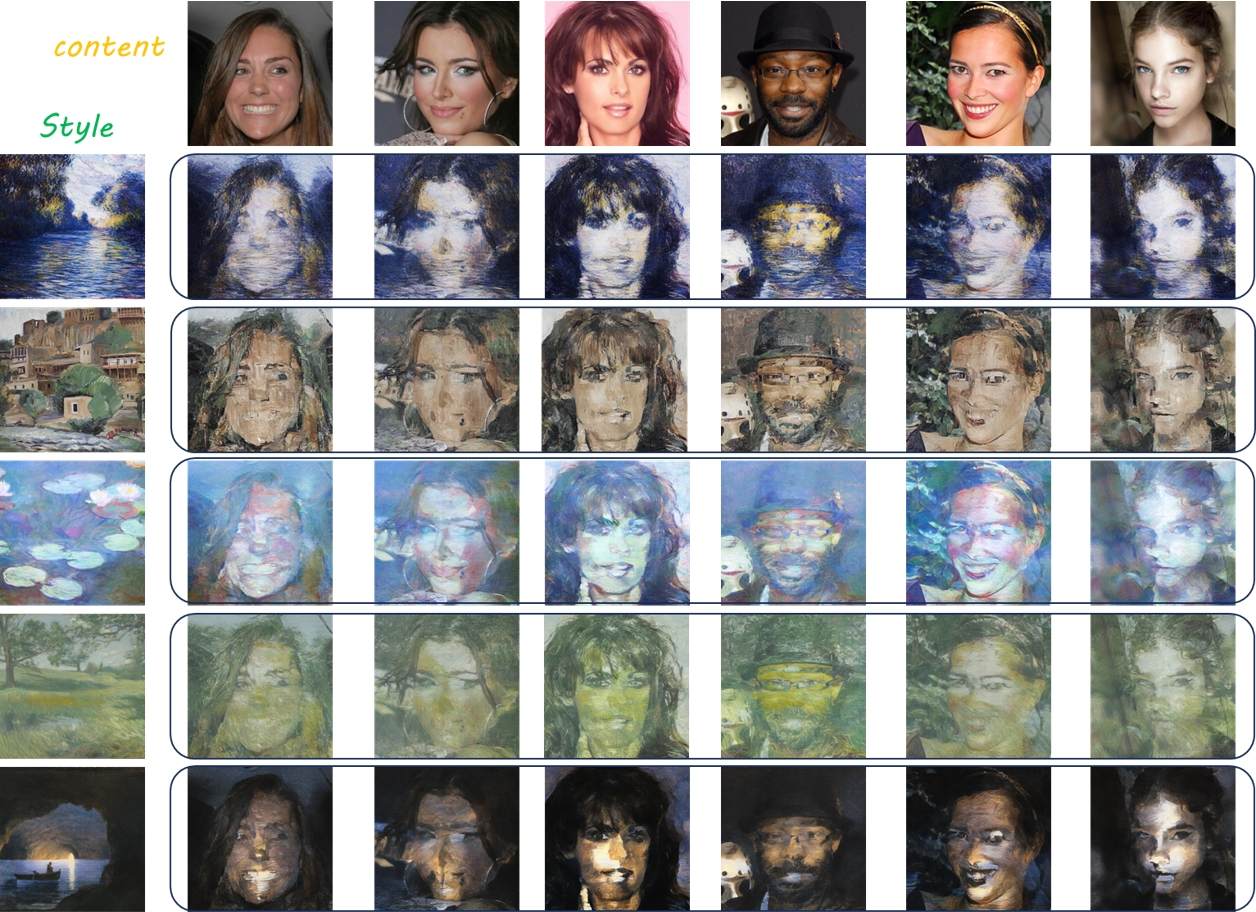}
     \caption{Additional style transfer results using non-facial artistic references. Each row shows a different example with content image, non-facial style reference, and our stylization result.}
    \label{fig:app_nonface}
\end{figure*}

We provide additional visualization results of style transfer from non-facial artistic references to realistic face images. As shown in Figure~\ref{fig:app_nonface}, our method successfully extracts and applies artistic elements from diverse non-portrait artworks while preserving facial identity and human features.
\subsection{Style Transfer Results Using Different Stable Diffusion Model Versions}
\begin{figure*}[!h]
\begin{minipage}{0.95\textwidth}
     \includegraphics[width=\textwidth]{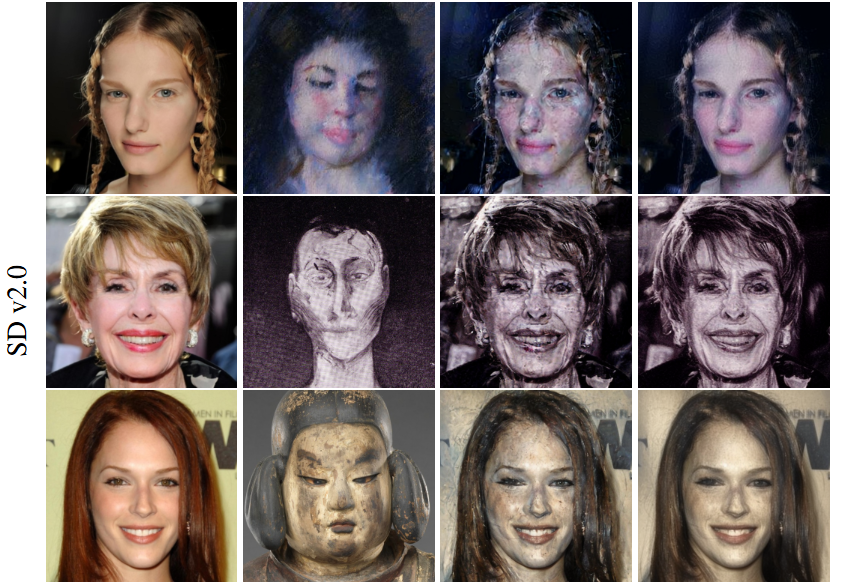}
    \end{minipage}
    
     \begin{minipage}{0.95\textwidth}
     \includegraphics[width=\textwidth]{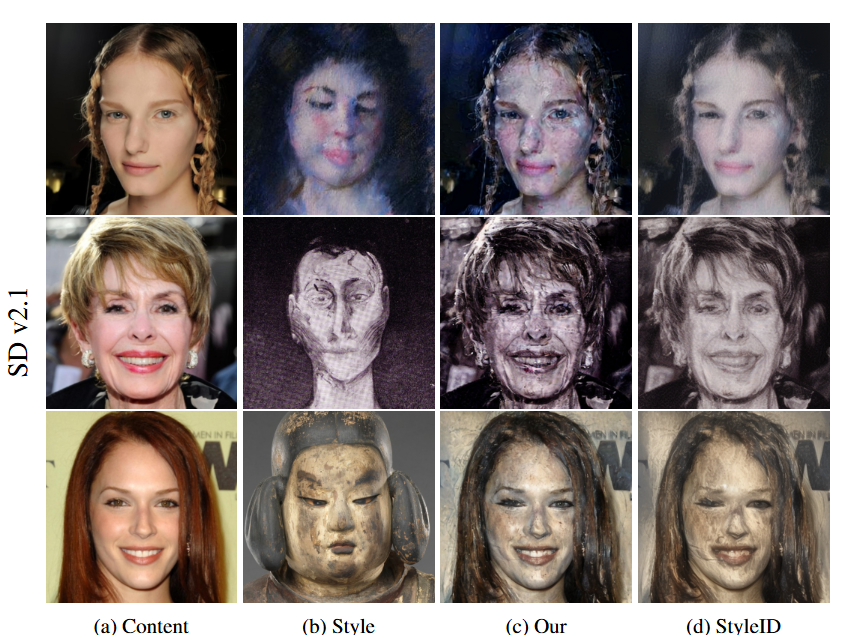}
    \end{minipage}
\caption{Style transfer results using different versions of Stable Diffusion models.}
\label{fig:sd_results1}
\end{figure*}

\begin{figure*}[!h]
\includegraphics[width=\linewidth]{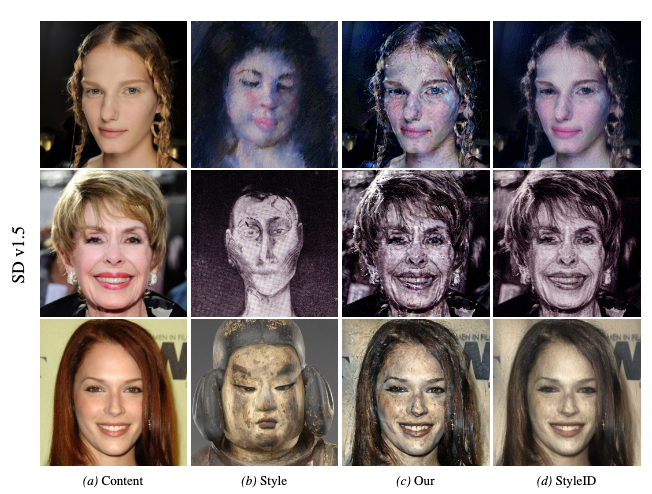}
\caption{Style transfer results using different versions of Stable Diffusion models.}
\label{fig:sd_results}
\end{figure*}

To evaluate the generalization capability and robustness of our proposed method, we conduct comprehensive experiments using multiple versions of Stable Diffusion models, specifically SD v1.5, v2.0, and v2.1. Figure~\ref{fig:sd_results} and Figure~\ref{fig:sd_results1} present a comparative analysis of style transfer results across these model variants. Our empirical findings demonstrate that the method exhibits consistent performance characteristics across different model architectures, maintaining robust facial identity preservation, semantic content fidelity, and effective artistic style element extraction and application. The consistent performance across model versions validates the architecture-agnostic nature of our approach, showing superior results compared to baseline methods, particularly StyleID. This characteristic is particularly valuable as it ensures reliable style transfer capabilities regardless of the underlying diffusion model implementation, contributing to the method's practical applicability and deployment flexibility.

%\input{checklist.tex}

%%%%%%%%%%%%%%%%%%%%%%%%%%%%%%%%%%%%%%%%%%%%%%%%%%%%%%%%%%%%%%%%%%%%%%%%%%%%%%%
%%%%%%%%%%%%%%%%%%%%%%%%%%%%%%%%%%%%%%%%%%%%%%%%%%%%%%%%%%%%%%%%%%%%%%%%%%%%%%%
% APPENDIX
%%%%%%%%%%%%%%%%%%%%%%%%%%%%%%%%%%%%%%%%%%%%%%%%%%%%%%%%%%%%%%%%%%%%%%%%%%%%%%%
%%%%%%%%%%%%%%%%%%%%%%%%%%%%%%%%%%%%%%%%%%%%%%%%%%%%%%%%%%%%%%%%%%%%%%%%%%%%%%%

\end{document}